\documentclass[sigconf]{acmart}
\usepackage{graphicx}
\usepackage{amsmath}
\usepackage{diagbox}
\usepackage{bbding}
\usepackage{booktabs}
\usepackage{hyperref}
\usepackage{makecell}
\usepackage{bm}
\usepackage{multirow}
\usepackage{balance}
\usepackage{tabu}
\usepackage{caption}
\usepackage{subcaption}

\AtBeginDocument{%
}

\setcopyright{acmcopyright}
\copyrightyear{2022} 
\acmYear{2022} 
\setcopyright{acmcopyright}\acmConference[MM '22]{Proceedings of the 30th ACM International Conference on Multimedia}{October 10--14, 2022}{Lisboa, Portugal}
\acmBooktitle{Proceedings of the 30th ACM International Conference on Multimedia (MM '22), October 10--14, 2022, Lisboa, Portugal}
\acmPrice{15.00}
\acmDOI{10.1145/3503161.3547953}
\acmISBN{978-1-4503-9203-7/22/10}

\begin{document}

\title{Parameterization of Cross-Token Relations with Relative Positional Encoding for Vision MLP}

\author{Zhicai Wang}
\email{wangzhic@mail.ustc.edu.cn}
\affiliation{%
  \institution{University of Science and Technology of China}
  \city{Hefei}
  \state{Anhui}
  \country{China}
  \postcode{230052}
}
\author{Yanbin Hao$^{\S}$}
\email{haoyanbin@hotmail.com}
\affiliation{%
  \institution{University of Science and Technology of China}
  \city{Hefei}
  \state{Anhui}
  \country{China}
  \postcode{230052}
}
% \thanks{$^\S$ Yanbin Hao is the corresponding author.}

\author{Xingyu Gao$^{\S}$}
\email{gxy9910@gmail.com}
\affiliation{%
  \institution{Institute of Microelectronics, Chinese Academy of Sciences}
  \city{Beijing}
  \country{China}
  \postcode{100045}
}

\author{Hao Zhang}
\email{zhanghaoinf@gmail.com}
\affiliation{%
  \institution{Singapore Management University}
  \country{Singapore}
  \postcode{230052}
}

\author{Shuo Wang}
\email{shuowang.hfut@gmail.com}
\affiliation{%
  \institution{University of Science and Technology of China}
  \city{Hefei}
  \state{Anhui}
  \country{China}
  \postcode{230052}
}

\author{Tingting Mu}
\email{tingtingmu@manchester.ac.uk}
\affiliation{%
  \institution{University of Manchester}
  \city{Manchester }
  \country{United Kingdom}
  \postcode{230052}
}

\author{Xiangnan He}
\email{xiangnanhe@gmail.com}
\affiliation{%
  \institution{University of Science and Technology of China}
  \city{Hefei}
  \state{Anhui}
  \country{China}
  \postcode{230052}
}

% \author{Zhicai Wang$^1$, Yanbin Hao$^{1\S}$, Xingyu Gao$^{2\S}$, Hao Zhang$^3$, Shuo Wang$^1$, Tingting Mu$^4$, Xiangnan He$^1$}
\thanks{$^\S$ Yanbin Hao and Xingyu Gao are both the corresponding author.}
% \email{wangzhic@mail.ustc.edu.cn, haoyanbin@hotmail.com, {gxy9910,zhanghaoinf}@gmail.com}
% \email{tingtingmu@manchester.ac.uk, xiangnanhe@gmail.com}
% \affiliation{%
%   \institution{$^1$University of Science and Technology of China, $^2$Institute of Microelectronics, Chinese Academy of Sciences, $^3$Singapore Management University, $^4$The University of Manchester}
%   \city{}
%   \country{}
% }
% \affiliation{
%   \institution{$^2$National University of Singapore}
% }
% \affiliation{%
%   \institution{wangzhic@mail.ustc.edu.cn, haoyanbin@hotmail.com}
%   \city{Hefei}
%   \contry{China}
% \author{Yi Tan$^1$, Yanbin Hao$^{1\S}$, Hao Zhang$^2$, Shuo Wang$^1$, Xiangnan He$^1$}
% \thanks{$^\S$ Yanbin Hao is the corresponding author.}

% \affiliation{
%   \institution{$^1$University of Science and Technology of China, $^2$Singapore Management University}
% }
% % \affiliation{
% %   \institution{$^2$National University of Singapore}
% % }
% \email{ty133@mail.ustc.edu.cn, haoyanbin@hotmail.com, hzhang@smu.edu.sg}
% \email{{shuowang.hfut,xiangnanhe}@gmail.com}
%%
\title{Parameterization of Cross-Token Relations with Relative Positional Encoding for Vision MLP}

% From Zhicai: We modified the abstract, please check this.
\begin{abstract}
  \par Vision multi-layer perceptrons (MLPs) have shown promising performance in computer vision tasks, and become the main competitor of CNNs and vision Transformers.  They use token-mixing layers to capture cross-token interactions, as opposed to the multi-head self-attention mechanism used by Transformers.  However, the heavily parameterized token-mixing layers naturally lack mechanisms to capture local information and multi-granular non-local relations, thus their discriminative power is restrained. To tackle this issue, we propose a new positional spacial gating unit (PoSGU). It exploits the attention formulations used in  the classical relative positional encoding (RPE),  to efficiently encode the cross-token relations for token mixing. It can successfully reduce the current quadratic parameter complexity $O(N^2)$ of vision MLPs to  $O(N)$ and $O(1)$. We experiment  with two RPE mechanisms, and further propose a group-wise extension to improve their expressive power with the accomplishment of multi-granular contexts. These then serve as the key building blocks  of a new type of vision MLP, referred to as PosMLP. We evaluate the effectiveness of the proposed approach by conducting thorough experiments, demonstrating an improved or comparable performance with reduced parameter complexity. For instance, for a model trained on ImageNet1K, we achieve a performance improvement  from 72.14\% to 74.02\% and a learnable parameter reduction from $19.4M$ to $18.2M$. Code could be found at  \href{https://github.com/Zhicaiwww/PosMLP}{https://github.com/Zhicaiwww/PosMLP}.
\vspace{-0.2cm}
\end{abstract}
 %However, the current designs of these token-mixing layers are heavily parameterized,  which makes the pre-training very demanding. 

  

\renewcommand{\shortauthors}{Zhicai Wang et al.}
\maketitle
\section{Introduction}

\begin{figure}[t]
  \small
  \centering
      \includegraphics[width=\linewidth]{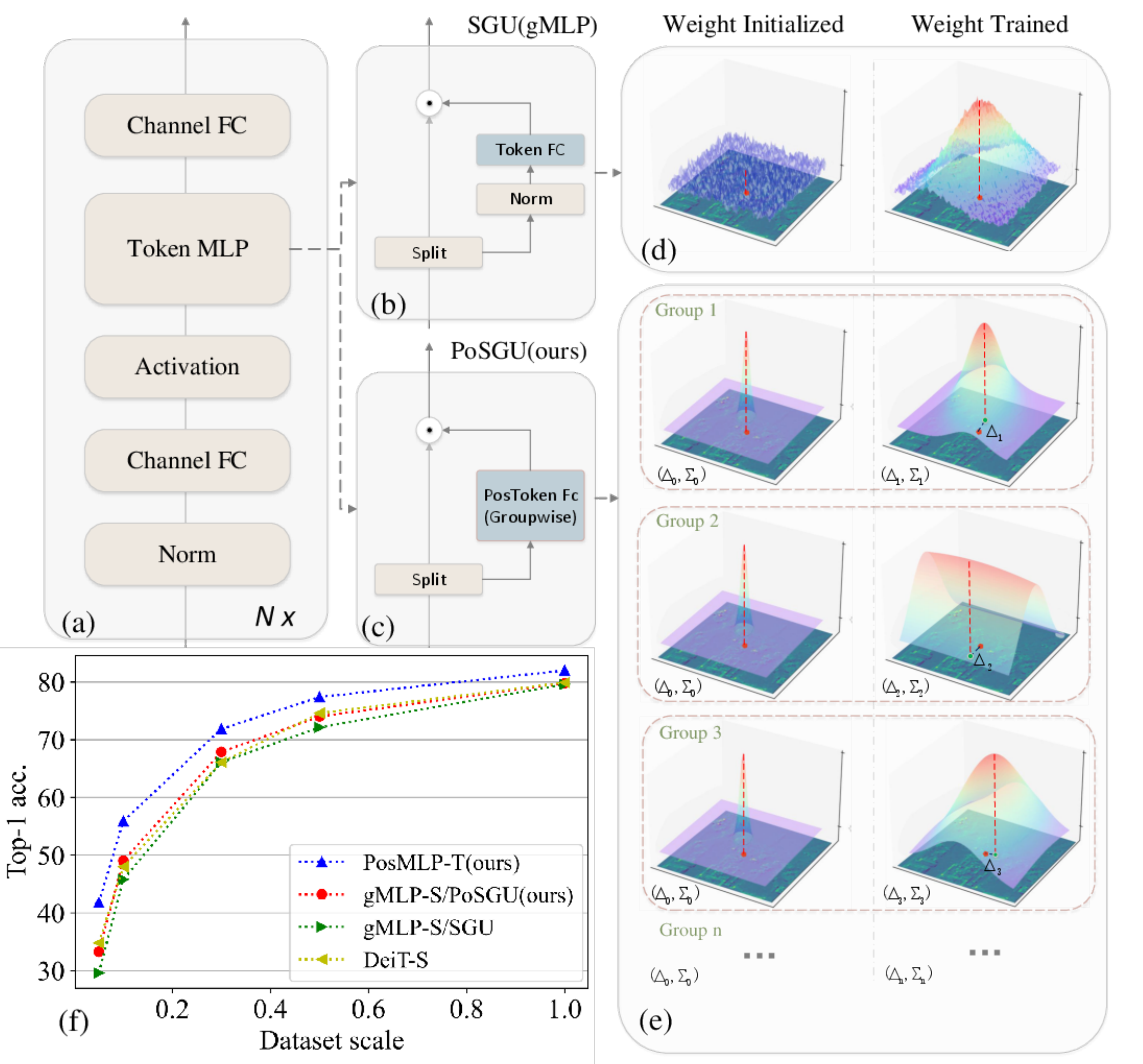} 
      \caption{(a), (b) and (c) present the basic network block,  the SGU used by gMLP \cite{Pay}, and the  proposed  PoSGU in a group-wise implementation,respectively. (d) and (e) visualize the projection weights of the SGU and PoSGU,respectively. Here, the red anchor represents a query token.  The rainbow distribution  shows the intensity of attention logits to its surrounding tokens. (f) shows the sample efficiency results for ImageNet2012 with 
      %From TT: What is sample ratio setting? Is it your dataset scale in figure?  Should keep consistency.
      modified sample ratio setting per class.}
      \vspace{-0.5cm}
      \label{fig:1}
  \end{figure}
In computer vision, the three  types of mainstream architecture CNN, Transformer and MLP support the backbone neural network design.
CNNs directly operate on 2D images, while vision transformers and MLPs convert each input image to a sequence of tokens.
Subsequently, they rely on different operational approaches to extract information from the images.
CNNs use convolutions to model local content patterns,  and use the weight sharing mechanism to improve design efficiency \cite{simoncelli2001natural,krizhevsky2012imagenet}. 
To capture cross-token interaction, vision transformers use the multi-head self-attention (MHSA) mechanism, and MLPs use the token-mixing layers.
As compared to CNNs,  the vision transformers and MLPs are notably better at their global receptive field, i.e. MHSA and token-mixing, and have the advantage of unsaturated discriminatory power when processing hyper-scale datasets \cite{tolstikhin2021mlp,dosovitskiy2020image}. 
%From TT: What is large-scale pre-training? Not clear. Require a lot of examples to pre-train, or need to pre-train a lot of parameters?  You talk a lot about MLP parameters later.  Shouldn't you mention parameters here?
%From Zhicai: Not modified. Here it should mean a lot of examples to pre-train. Parameter efficiency is a key point I want to highlight but the introduction of RPE do ease the pre-train scale. see the Fig 1 (f) that we have a better performance under same dataset scale.
But  they  bear the weakness of requiring large-scale pre-training, e.g.,  having more model parameters and thus consuming more training examples,  due to the weak inductive bias  \cite{touvron2021training,tolstikhin2021mlp,ding2021repmlp}. 
Various research effort has  been made to address this limitation for vision transformers, resulting in a series of new transformer design, e.g., locality-enhanced transformers (LocalViT) \cite{hu2019local}, Swin transformer \cite{liu2021swin} and NesT transformer \cite{zhang2021aggregating}.

%From TT: Is this the limitation?  I added parameterisation...
%From Zhicai: Not modified. The sample efficiency problem, i.e, the performance under samll dataset scale is typically limited
However, there is a research gap in vision MLP development, where the latest designs are still demanding in parameterization and pre-training.
For instance, to improve over MLP-Mixer \cite{tolstikhin2021mlp} that is the first MLP-based vision model,  
%From TT: You should mention spacial gating unit (SGU) here, to prepare for later.
%From Zhicai: Not modified.
gMLP \cite{Pay} uses a gating mechanism and a single-token fully connected (FC) layer, referred to as the spacial gating unit (SGU), to achieve efficient aggregation of spacial information. 
But it results in a dramatically increased parameter number as the token number grows larger.  
More recently, building on the idea of token-free substitute, axial shifted MLP (AS-MLP) \cite{lian2021mlp}, Permute-MLP \cite{hou2021vision}, Hire-MLP~\cite{guo2021hire} and spatial shift MLP (s$^{2}$-MLP) \cite{yu2021s} design different spatial shift variants to simulate token interaction.
As a result, they have managed to remove parameters depending on token dimensions but  created extra FC layers instead. 
This has actually increased their parameter numbers. 
%From TT: You don't need too much detail in introduction. It makes the reader tired. The goal is to only to highlight research gap you want to address.
%From Zhicai: Not modified but added. Additionaly add some supplementary statement here.
Additionally, their spatial shift replacement has caused a loss of intrinsic property when aggregating the global features in a token FC layer, i.e., explicit shift operation should gain global receptive field by deepening the network.

%From TT: It is better to propose one thing, including different design components. Your method proposed many different things, not focused...This results in a not very focused contribution paragraph.
%From Zhicai : Not modified. I would like to emphasize the proposition of 'PoSGU' rather than the backbone 'PosMLP', since this the main contribution for vision MLP models from my point.

%From TT: put a reference on RPE
%From Zhicai: Not modified but added.

To fill in this research gap, we propose to utilize the relative positional encoding (RPE) \cite{shaw2018self} to improve the vision MLPs, aiming at reducing the model parameterization  but without sacrificing the model expressive power. Our enhanced MLP is named as PosMLP. 
%From TT: Add result reference from your own secc
%From Zhicai: Not modified.
The  design is motivated by the fact that the ultimate goal of a token-mixing MLP is to model the relations across tokens (spatial pixels), \textbf{\textit{why not directly parameterize the relations in an efficient way}}.  
%From TT: Is this proposed by you?
%From Zhicai: modified. PoSGU is a modified version of SGU that is proposed by us. That's what I want to state. Mentioned PosToken FC and changed same statement, i.e, multi-granular is guaranteed by group-wise operation
Therefore, we make the following contributions:
\begin{itemize}
%From TT: Is this proposed by you?
%From Zhicai: modified. PoSGU is a modified version of SGU that is proposed by us. That's what I want to state. Mentioned PosToken FC and changed same statement, i.e, multi-granular is guaranteed by groupwise operation
%From TT: Check this. I polished the sentence. Is it still what you want to express?
\item We propose a positional spacial gating unit (PoSGU) to model cross-token interactions. Its main building block is a PosToken FC layer developed based on RPE that has been proven to be efficient to encode the interaction between the key and query items in attention modelling. 
\item We implement two RPE mechanisms in the PoSGU.  As compared to the quadratic parameter complexity $O(N^2)$ of gMLP, the first mechanism can reduce the complexity to $O(N)$ while the later to $O(1)$. 
\item Additionally, we propose a group-wise extension to achieve the multi-granular spacial feature aggregation and enhance the expressive power of the RPE. 
\end{itemize}
%From Zhicai:Added. Fig 1 I want to show readers the clear comparison between SGU and PoSGU, so I moved here and added some deatailed description of Fig 1
Figure \ref{fig:1} highlights the architecture difference between PoSGU (the GGQPE version by default, see Sec \ref{sec:4.1}) and SGU from the gMLP that both are sharing the same basic network block design. Subgraphs (d) and (e) show that our token-mixing logits initialized in PoSGU are highly concentrated around the query token and the cross-token correlation is modeled by learned non-isotropic Gaussian distribution.  
%
%From TT: Question: You should mention why PosMLP has the advantage of easing large-scale training.  It seems your research gap to address.
%From Zhicai: Not modified. I could mention  the point but suffer from a lack of experimental results under large-scale training. The conclusion might be that the introduction of positional inductive bias will ease the training, while the group-wise operation will enrich the feature captured.    
%From TT: You can report some actual performance here, rather than keep talking the deasign advantage.
%From Zhicai: Modified. Mentioned the experimental figure in Fig 1 and put a comparison that shows how helpful of introducing RPE to gMLP. 
After a thorough evaluation,  the results show that a simple PoSGU  can result in satisfactory performance with a
significantly reduced parameter number, e.g. a comparable performance to the transformer-based DeiT-S \cite{touvron2021training}  as in Figure \ref{fig:1} (f). For instance, for a model trained on ImageNet1K with 
%From TT: what is a 0.5 sample ratio? Can you  explain?
0.5 sample ratio,  our approach can achieve a performance improvement  from 72.14\% to 74.02\% and a parameter (learnable) reduction from $19.4M$ to $18.2M$.

\section{Related Work}

\textbf{Vision transformers.} In recent years, vision transformers have shown promising image classification results for his strong power on modeling long-range dependencies over pixels \cite{liu2021swin,zhang2021token,fang2022cross,cheng2018intrinsic}. The vision transformer (ViT) \cite{dosovitskiy2020image} is the pilot work that extends the linguistical-style transformers on visual content modeling by treating a 2D image as a sequence of 1D tokens (patches) while remaining the vision robustness \cite{bhojanapalli2021understanding}. DeiT\cite{touvron2021training} further adopts distillation \cite{hinton2015distilling} from a CNN teacher and extensive data augmentations to successfully boost the image classification performance. To be friendly to more downstream tasks like object detection and segmentation \cite{lin2014microsoft}, PVT~\cite{wang2021pyramid} provides the hierarchical architecture for efficient vision transformer. Swin transformer \cite{liu2021swin} further manipulates shifted windows into the hierarchical architecture and gains notable performance improvement. To further enhance with local information, CoAtNet \cite{dai2021coatnet} explicitly integrates convolutional blocks into the feedforward transformer network.

\textbf{Vision MLPs.} MLP-like models were also developed under the premise of large-scale pre-training\cite{tolstikhin2021mlp,touvron2021resmlp}. Equipped with normalization \cite{ba2016layer} and residual connection \cite{he2016deep}, the architecture of stacking pure linear layers can also achieve comparable results with CNNs. However, the token mixing MLP in MLP-Mixer \cite{tolstikhin2021mlp} and gMLP \cite{Pay} introduce few inductive biases and constraint the input resolution. To overcome these, bunch of works are built upon the consumption-free shift\&rearrange operation, e.g. 
s$^{2}$-MLP \cite{yu2021s} that uses spatial shifing MLP to replace token mixing MLP, AS-MLP\cite{lian2021mlp} that axially shift feature in the spacial dimension, Vision Permutator \cite{hou2021vision} that uses Permute-MLP to mix height and width information simultaneously, and Hire-MLP \cite{guo2021hire} that uses the pairwise rearrange\&restore  manipulations to realize local and global feature exchanging over patches. Besides, Rep-MLP \cite{ding2021repmlp} implemented reparameterization to merge trained convolution layers into a pure MLP network for evaluation. Different from these aforementioned methods,
our PosMLP first implements positional encoding as a fully prior to explicitly model token interactions in MLP, and hence achieves better sample and parameter efficiency.   

\textbf{Positional Encoding.} Transformer is inherently lack of sensitivity to token position, and so is MLPs. Positional encoding (PE) is a fairly mature method implemented in transformer to overcome this problem \cite{radford2019language,radford2018improving,ju2020transformer}. There are commonly two kinds of positional encoding (PE) methods, i.e., absolute PE (APE) and relative PE (RPE). Sinusoidal positional encoding is the first APE implemented in vanilla transformer \cite{Attention}, aiming to introduce a prior of absolute token position distinction. RPE was first introduced to amplify the natural sensitivity of relative distance in linguistic expressions\cite{shaw2018self,he2020deberta,huang2020improve} and late extensively adopted in vision transformers \cite{Rethinking} which requires a strong prior of 2D displacement between pixels. Original RPE is based on a learnable relative positional embedding and thus introduces a soft positional inductive bias. CAPE \cite{likhomanenko2021cape} utilizes an enhaced APE strategy to realize the model generalization as introduced by RPE. By contrast, Quadratic Positional Encoding (QPE) \cite{Onthe} was proposed to bring a pre-defined relative position embedding to multi-head self attention (MHSA) such that it can act like a convolutional layer. ConViT \cite{Convit} integrates QPE into DeiT \cite{touvron2021training} and achieves a higher sample efficiency. Our work further develops Generalized QPE into a group-wise variant for vision MLPs, such that it obtains better expression ability.

\section{Preliminaries}

In this section, we first introduce the design of spacial gating unit (SGU) of gMLP. Next, we summarize a general representation for two representative relative positional encoding (RPE) methods, i.e., learnable relative positional encoding (LRPE) and generalized quadratic positional encoding (GQPE), which are commonly used by vision Transformers. 
% In this work, we instantiate two representative RPE methods to design our vision MLP.
\subsection{Spatial Gating Unit } 
%
%From  TT: I don't suggest present too much result in introduction and method section... 
%From TT: I suggest to move this figure to the SGU section.
%From Zhicai: Not modified. This suggestion is helpful, and this make the paper easier to read 

SGU is the main building block of gMLP \cite{Pay}, used to enable cross-token interactions. 
Let $\mathbf{X}\in \mathbb{R}^{N\times d}$ be the token representations, 
where $N$ and $d$ denotes the token number  and embedding (channel) dimension, respectively. 
%From TT: You refer d as embedding and channel dimension.  Which one is right? Use one.
%From Zhicai: Not modified. 'd' should represent channel dimension and I will try to guarantee this in the paper.
SGU splits the token representations into two independent parts along the dimension, 
denoted by $\mathbf{X}^{1}\in \mathbb{R}^{N\times \frac{d}{2}}$ and $\mathbf{X}^{2}\in \mathbb{R}^{N\times \frac{d}{2}}$. 
It uses a linear projection on one part to calculate a gating mask, then uses the mask to refine element-wisely the other part. 
Denote the refined token representations by $\mathbf{Z}\in \mathbb{R}^{N\times \frac{d}{2}}$, and it is     computed  by
\begin{equation}
\mathbf{Z} = \left (  \mathbf{W} \textmd{Norm}(\mathbf{X}^{1})+\bm b \right ) \odot \mathbf{X}^{2} ,
\label{eq:1}
\end{equation}
where  $\mathbf{W}\in \mathbb{R}^{N\times N}$, $\bm b \in \mathbb{R}^{N}$  and $\odot$ denotes 
the Hadamard product. $\textmd{Norm}(\cdot)$ here is the $LayerNorm$ \cite{ba2016layer}. 
The projection matrix $\mathbf{W}$ is for feature refinement, and the gating operation encourages higher-order information interactions \cite{Pay}.
%From TT: You should mention the concept "token FC" here and refer to your Figure 1(b). I guessed the following. Check.
%From Zhicai: Modified. Token FC should represent Liner mapping sololy and Hadamard product represents gating operation. 
As shown in the SGU diagram in  Figure \ref{fig:1}(b), the operation $\tilde{\mathbf{X}}^{1}=\textmd{Norm}(\mathbf{X}^{1})$ is referred to as the Norm layer, and the operation of $\left ( \mathbf{W} \tilde{\mathbf{X}}^{1} +\bm b \right ) $ is referred to as the Token FC layer. Specifically, the Hadamard product is representing as the gating operation.
\vspace*{-0.2cm}

\subsection{Relative Positional Encoding} 
%From TT: Some reference would be good?
%From Zhicai: Not modified but added. I have mentioned in related work, however I addtionly put some here.
Both absolute and relative positional encoding mechanisms have been widely used in vision transformers \cite{Rethinking,chu2021conditional,dong2021cswin}. We introduce only RPE since our research focus is to model the pairwise relations between tokens. 
The core idea of RPE is to enhance the calculation of the attention $\mathbf{A}\in \mathbb{R}^{N\times N}$ by considering the position difference between tokens. 
 In general, the token position is marked in a square window\footnote{In vision transformers, e.g., Swin transformer \cite{liu2021swin}, a square window ($\sqrt{N}\times \sqrt{N}$=$N$ tokens) is commonly used.} $\left(\sqrt{N}, \sqrt{N} \right)$ and therefore $\bm \delta$ lies in the range $\left[-\sqrt{N}+1,\sqrt{N}-1\right]\times\left[-\sqrt{N}+1,\sqrt{N}-1\right]$.
A relative position is a $2$-dimensional   vector  $\bm \delta_{i,j} =\bm p_{j}-\bm p_{i} $ where $\bm p_{i},\bm p_{j}$ are the positions of the $i$th and $j$th tokens in the square window.
% From TT: I don't understand this.
% From Zhicai: Not modified. Typically there are similar group operation in transformer, but I don't want mix it up with ours.
For simplicity, we omit the head dimension (i.e., single-head). 
The pipeline formulation of RPE can be generalized as 
\vspace{-0.25cm}
\begin{align}
\mathbf{A}_{i,j} :=&\; \textmd{Softmax}_{j}\left(\mathbf{W}_{i,j} + \mathbf{W}^{r}_{i,j}\right),\\
\label{eq:2}
\mathbf{W}^{r}_{i,j} :=&\; \bm v \bm r^\top_{\bm \delta_{i,j}},\\
\label{eq:Wrij}
\mathbf{W}_{i,j}:=&\; \mathbf{Q}_{i,:}\mathbf{K}^\top_{j,:},
\end{align}
where $1\leq i,j\leq N$, $\mathbf{Q}$ and $\mathbf{K}$ are the query and key matrices in a transformer, and $\bm v \in \mathbb{R}^{D_{pos}}$ is a projection vector that aggregates the relative positional embedding $\bm r_{\bm \delta} \in \mathbb{R}^{D_{pos}}$ to obtain the attention bias. 
Different RPE approaches vary in the formulation of $\bm r_{\bm \delta}$ and $\mathbf{W}_{i,j}^{r}$.  The two representative RPE algorithms  of learnable relative positional encoding (LRPE) \cite{hu2018relation,he2017mask,raffel2019exploring} and generalized quadratic positional encoding (GQPE) \cite{Onthe} are used in our work.

{\bf Learnable Relative Positional Encoding:  }  LRPE simply sets $D_{pos}=1$. Consequently, both $\bm v$ and $\bm r$ in $\mathbf{W}^{r}$ collapse to a single scalar and $v$ is further set to constant 1. It constructs a learnable relative positional
%From TT: Check notation. I added $\bm \delta$ as subscript.
%From Zhicai: Modified. Make the LRPE case clear  
embedding dictionary $\bm r_{\bm \delta}^{\textmd{lrpe}} \in \mathbb{R}^{\left(2\sqrt{N}-1\right)^2}$ with a parameter complexity of $\mathcal{O}(N)$. Here,  $\bm \delta$ is used as an index. 

{\bf Generalized Quadratic Positional Encoding:  }    GQPE was firstly proposed  for MHSA.  It mimics the function of the convolutional layer by forming a non-isotropic Gaussian distribution over the query token:
\vspace{-0.2cm}
\begin{align}
\vspace{-0.1cm}
    \mathbf{A}^{\textmd{gqpe}}_{i,j} &:= \textmd{Softmax}_j\left(-\frac{1}{2}\left(\bm \delta_{i,j} - \bm \Delta\right)\mathbf{\Sigma}^{-1}\left(\bm \delta_{i,j} - \bm \Delta\right)^\top\right),\\
    &= \textmd{Softmax}_j\left(\bm v \bm r^\top_{\bm \delta_{i,j}}\right).
    \label{eq:5}
\end{align}
To obtain such a distribution pattern,  the learnable vector $\bm v$ is parameterized by a center of attention $\mathbf{\Delta}\in \mathbb{R}^2$ and a positive semi-definite covariance matrix $\mathbf{\Sigma} = \mathbf{\Gamma}  \mathbf{\Gamma} ^\top $, where $ \mathbf{\Gamma}  \in\mathbb{R}^{2\times2}$. In total, there are 6 parameters.  The parameter complexity is $\mathcal{O}(1)$. Subsequently, 
\begin{align}
 \label{eq:6}
    \bm v^{\textmd{gqpe}}= & \left[\left(\mathbf{\Sigma}^{-1}\mathbf{\Delta}\right)_1, \left(\mathbf{\Sigma}^{-1}\mathbf{\Delta}\right)_2, -\frac{1}{2}\mathbf{\Sigma}_{1,1}^{-1}, -\frac{1}{2}\mathbf{\Sigma}^{-1}_{2,2}, -\mathbf{\Sigma}^{-1}_{1,2}\right]^\top, \\
\label{eq:7}
\bm r^{\textmd{gqpe}}_{\bm \delta}=  & \left[\bm \delta_x,\bm \delta_y,\bm \delta^2_x,\bm \delta^2_y,\bm \delta_x \bm \delta_y\right]^\top.
\end{align}
The relative positions $\bm r_{\bm \delta}$ of GQPE is not learnable and is determined only by the relative position $\bm \delta = (\bm \delta_x, \bm \delta_y)$. 
For the attention map of the $i$-th query token, $\bm \Delta$ controls the displacement of the distribution center relative to the position $\bm p_i$, and $\mathbf{\Sigma}$ controls the distribution function where it is semi-definite positive such that the attention maximum will always be guaranteed at $\bm \Delta = \bm \delta$. 

%From TT: Why do you need this statement?  Is this concept "hard" important for claiming your contribution?
%From Zhicai: Not modified. I tended to emphasize the difference between LRPE and GQPE  

Since GQPE introduces a hard relative position prior,  we view it as a \textbf{hard} RPE similar to the hard inductive bias introduced by CNN \cite{Convit}.

\section{ Positional MLP (PosMLP)}
\label{sec:4}

%From TT: When you refer to token FC layer, do you mean SGU? Is this used by gMLP only or many vision MLPs. Check the text below. I am not sure whether I understood you correctly.
%From Zhicai: Not modified. Token FC sololy means the original linear projection matrix here

The Token FC layer used in classical vision  MLPs bears the weakness of poor inductive biases \cite{tolstikhin2021mlp} and has a quadratic parameter complexity $\mathcal{O}(N^2)$ with respect to the number of tokens. We propose to transfer the idea of RPE used in attention formulation for transformers to formulate the projection matrix in SGU. This introduces extra positional information in the Token FC layer of SGU. Therefore, we  name the improved SGU as PoSGU, and the improved vision MLP with such new units as PosMLP. This proposal can successfully reduce the parameter complexity of token-mixing to only $\mathcal{O}(N)$ or $\mathcal{O}(1)$ without sacrificing the model performance.

\begin{figure*}[h]
  \centering
  \includegraphics[width=1\linewidth]{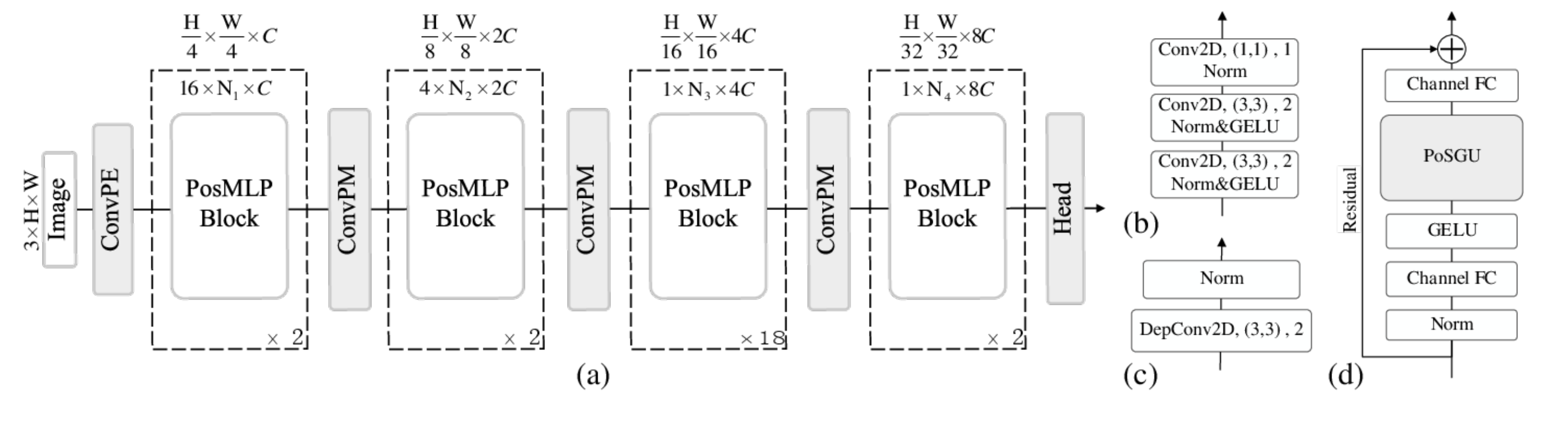}
  \vspace{-0.9cm}
  \caption{\textbf{The proposed PosMLP:} (a) Overall architecture; (b) Convolutional Patch Embedding block; (c) Convolutional Patch Merging block; (d) Architecture of PosMLP block with PoSGU.}
  \vspace{-0.3cm}
  \label{fig:PosMLP}
\end{figure*}

\subsection{Positional SGU (PoSGU)}
\label{sec:4.1}
Starting from the classical SGU  operation  $\left ( \mathbf{W} \tilde{\mathbf{X}}^{1} +\bm b \right ) \odot \mathbf{X}^{2}$ as explained in preliminaries, we  omit $\bm b$ since the bias term $\bm b$  operates independently on $\mathbf{X}^{2}$.  We will also discuss this in the experiment later. This results in $\left ( \mathbf{W} \tilde{\mathbf{X}}^{1} \right ) \odot \mathbf{X}^{2}$, based on which we propose the  base formulation of PoSGU, as
\vspace{-0.1cm}
\begin{equation}
\vspace{-0.1cm}
\mathbf{Z}=\left ( \left(\mathbf{W} +\mathbf{W}^{r} \right ) \tilde{\mathbf{X}}^{1} \right)\odot \mathbf{X}^{2},
\end{equation} 
where $ \mathbf{W}^r $ encodes the positional information and is formulated by taking advantage of the attention weight formulation in Eq. (\ref{eq:2}).    

%
%From TT:  It does not seem right here. r is a vector, W is a matrix.  how can you add them?
%From Zhicai:Modified, orporate---->cooperate. This suppose to be right. r is a matrix here, since the idex $\delta$ is a matrix and it plays a role of re-index. 
{\bf  LRPE-based PoSGU:} When following the LRPE principal,  $\mathbf{W} $ can be simply modified to $\mathbf{W}+\bm r^{\textmd{lrpe}}_{\bm\delta}$ in order to cooperate the positional information. This results in 
 \begin{equation}
    \mathbf{Z}^{\textmd{lrpe-M}} =\left (\mathbf{W}+ \bm r^{\textmd{lrpe}}_{\bm\delta} \right)\textmd{Norm}(\mathbf{X}^{1})\odot \mathbf{X}^{2},
      \label{eq:9}
    \end{equation}
%From  TT: Wehre is $\mathbf{W}^{r}$?
%From Zhicai:$$\mathbf{W}^{r} = \bm r^{\textmd{lrpe}}_{\bm\delta}$ in LRPE case, since v is a constant of 1.
where the relative positional encoding matrix $\mathbf{W}^{r}$ can be calculated by $\bm r^{\textmd{lrpe}}_{\bm\delta}$.
It is worth to mention that we remove the \textmd{Softmax} operation, due to its co-occurrence with the \textmd{LayerNorm} operation\footnote{The \textmd{Norm} operation here is used to enable a direct comparison with gMLP.} will cause a significant performance drop. Note that our LRPE-M shares a similar spirit with the bias mode of the iRPE \cite{wu2021rethinking}.
Through empirical observations,  e.g., in the
%From TT:  I don't understand... Need more context. What is a query token, related to i?
%From Zhicai: Modified. I put some more details on appendix. Query token is the reference point, since the projection map is NxN and we need to choose a reference point with size 1xN that visualized in a \sqrt{N}x\sqrt{N} map.
Figure S\ref{fig:8} where we visualized each projection matrix and plotted the 
logits intensity corresponding to the query token.
We notice that the positional embedding $\bm r^{\textmd{lrpe}}_{\bm\delta}$  can learn more localities and can reserve  partially  its non-locality in deeper layers, while the effect of $\textbf{W}$ is weak and sometimes it can be redundant.  Therefore, we remove $\textbf{W}$ and this results in 
\begin{equation}
\mathbf{Z}^{\textmd{lrpe}} = \bm r^{\textmd{lrpe}}_{\bm\delta} \textmd{Norm}(\mathbf{X}^{1}) \odot \mathbf{X}^{2}.
  \label{eq:10}
 \end{equation}
%  \begin{equation}
%   \mathbf{Z}^{\textmd{lrpe}} = (\bm r^{\textmd{lrpe}}_{\bm\delta} + \bm b) \textmd{Norm}(\mathbf{X}^{1}) \odot \mathbf{X}^{2}.
%     \label{eq:10}
%    \end{equation}
 More importantly, this revised operation significantly decreases the parameter complexity.

%From TT: Check my early comments on embedding dimension and channel dimension. Also check the changed English. I changed it to match your maths. But I could be wrong.
%From Zhicai: Both description should be correct.

{\bf  GQPE-based Group-wise PoSGU:} An alternative is to follow the GQPE principal. The particular benefit is its extremely low parameter complexity $\mathcal{O}(1)$. Meanwhile, to increase the model expressive power, we enhance it with a group-wise strategy \cite{hao2022group}. Inspired by the multi-head operation used by transformers and the group convolution  operation used by  CNNs, we find it 
beneficial to split further the token embeddings along their channel dimension, and then  project different splits with  different  learnable vectors. 
Specifically, after splitting the token representations $\mathbf{X}\in \mathbb{R}^{N\times d}$ into  $\mathbf{X}^{1} \in \mathbb{R}^{N\times \frac{d}{2}}$  and $\mathbf{X}^{2} \in \mathbb{R}^{N\times \frac{d}{2}}$, we further split each into $s$ feature groups: $\left \{ \mathbf{X}^{1}_{1}, \mathbf{X}^{1}_{2}, \cdots,\mathbf{X}^{1}_{s}\right \} \in \mathbb{R}^{N\times \frac{d}{2s}}$ and $\left \{ \mathbf{X}^{2}_{1}, \mathbf{X}^{2}_{2}, \cdots,\mathbf{X}^{2}_{s}\right \} \in \mathbb{R}^{N\times \frac{d}{2s}}$.  
Following the idea of Eq. ( \ref{eq:5}), we  have, for the $s$-th group,
\begin{equation}
    \mathbf{Z}^{\textmd{gqpe}}_s = \mathbf{A}^{\textmd{gqpe}}_s\mathbf{X}^{1}_{s}\odot \mathbf{X}^{2}_{s},
    \label{eq:11}
  \end{equation}
% \begin{equation}
%     \mathbf{Z}^{\textmd{gqpe}} = (\mathbf{A}^{\textmd{gqpe}} + \bm b)\mathbf{X}^{1}\odot \mathbf{X}^{2},
%     \label{eq:11}
%   \end{equation}
\begin{equation}
    \mathbf{A}^{\textmd{gqpe}}_{s,i,j} = \textmd{Softmax}_j\left( \bm v^{\textmd{gqpe}}_s \bm r^{\textmd{gqpe}^{\top}}_{\bm \delta_{i,j}}\right).
      \label{eq:ggqpe}
\end{equation}
Then, we concatenate all the group representations:   		
 \begin{equation}
    \mathbf{Z}^{\textmd{gqpe}}=\textmd{Concat}\left(\mathbf{Z}^{\textmd{gqpe}}_2, \mathbf{Z}^{\textmd{gqpe}}_2, \cdots, \mathbf{Z}^{\textmd{gqpe}}_s\right).
      \label{eq:13}
\end{equation}
\vspace{-0.1cm} 
The $s$ groups share the same relative positional embedding $\bm r^{\textmd{gqpe}}_{\bm \delta}$ but  different learnable vector $\bm v^{\textmd{gqpe}}_s$. We omit the Norm layer here since the co-occurrence of Norm and Softmax will cause a performance drop (check Section \ref{sec:5.2})
The calculation builds on multiple pairwise shift attention centers $\{\mathbf{\Delta}_1,\mathbf{\Delta}_2,\cdots,\mathbf{\Delta}_s\}$ and covariance matrices $\{\mathbf{\Sigma}_1,\mathbf{\Sigma}_2,\cdots,\mathbf{\Sigma}_s\}$. A \{shift attention center, covariance matix\} pair can lead to a particular granularity of contextual information (local or non-local, mostly determined by the form of $\mathbf{\Sigma}$ as Sec.\ref{sec:4.2} shows). 
Therefore, this group-wise operation can improve over the original GQPE, being able to flexibly attend multiple granularities of contextual information in a single layer. 
%From TT: Why do you need to include a result Table in method section? Sometimes, it can be a distraction to talk about  result in method section... If you really have to do it, need to have clarity and also be sharp in your conclusion drawn from results.   I feel the quality of your text for presenting result in method section is higher than in experiment section.
%From Zhicai: Modified below. I was tend to emphasize the effectiveness of introduing RPE into MLP rather than I proposed a new backbone variant PosMLP. So I wanted to highlight this contribution on method part. Actually I have only tested PosMLP/GGQPE version on experiment, thus I need to make the transition from LRPE to GQPE be more resonable.
%From TT: Do you have a more specific subsection number to refer to here?
It  can potentially enrich the captured spatial information and ease the training (see Section \ref{sec:5.1} for results). 
%From TT: check here. Do you need some details, e.g., in footnote?
It is worth to mention that this group-wise strategy can be easily transferred to the LRPE mechanism, resulting in the GLRPE.
We provide a simple illustration in Figure S\ref{fig:7}, to highlight the difference between the classical Token FC  and the proposed token-mixing methods based on LRPE and GQPE.PE and GQPE-based token-mixing methods in this section, see Figure S\ref{fig:7}.

\vspace{-0.1cm}
\subsection{PosMLP Architecture}
\label{section:3.3}
We propose a new vision MLP model PosMLP, of which each basic block is integrated with our proposed PoSGU.
For illustration purpose, Figure \ref{fig:PosMLP} depicts the model architecture of a tiny PosMLP, which is mainly composed of the convolutional downsampling (ConvPE + ConvPM) and PosMLP block. 

{\bf Convolutional Patch Embedding (ConvPE):} The ConvPE block receives the input of a raw image with a shape $H\times W\times 3$ and outputs the patch embeddings. The module
shares the similar spirit as \cite{xiao2021early} in transformers, which stacks several convolutional layers to improve the stability of the model training. In our implementation, as shown in Figure \ref{fig:PosMLP}(b),  the block contains three consecutive convolutional layers. The use of convolutions adequately encodes the local information in visual modelling. Finally, this operation results in a $\frac{H}{4}\times \frac{W}{4}\times C$ feature map.

{\bf Convolutional Patch Merging (ConvPM):}
Inspired by  Nest Transformer \cite{zhang2021aggregating} that uses the ``Convolution+Maxpooling'' to perform the spacial downsampling with the ratio of $2$ and a channel expansion ratio of $2$, we adopt a depth-wise convolution with stride 2 to achieve a similar effect. 

{\bf Positional encoding MLP:} The architecture of a PosMLP block is illustrated in Figure \ref{fig:PosMLP} (d). It has a similar structure to gMLP, but replaces the SGU with the proposed PoSGU. Thanks to the RPE mechanisms, it is able to model both local and non-local information, thus can entirely replace the token-mixing operations in the original vision MLPs. 

{\bf Windows Partitioning:} To be fed into a PosMLP block, the image input is blocked into multiple non-overlapping windows of the given size and sequenced as tokens. Then it is reshaped back to a feature map for performing convolutional downsampling. In our implementation, we assign different window sizes to different stages by considering the feature map size, i.e., $14\times 14$ for the first three stages ``Stage 1,2,3'', resulting in the  $14\times 14=196$ (i.e., $N=196$) tokens (pixels) and the corresponding $16,4,1$ feature windows respectively, and $7\times 7$ for the ``Stage 4'' and $49$ tokens with $1$ feature windows. The most obvious advantage of using a windows-based structure is that it can significantly reduce the computational cost \cite{liu2021swin} and realize parameters sharing among windows.

{\bf Architecture Variant:}
We build our model variants similar to the classic works like Swin-Transformer, including three 4-stages PosMLP-T/S/B where the capital letters ``T, S, B'' refer to ``Tiny, Small, Base'' models. The main difference between PosMLP-T, PosMLP-S, and PosMLP-B lies in the model size which is led by separately setting the feature channel $C$ to 96, 128, and 192. All the other hyper-parameters such as window size, group numbrt $s$, and channel expansion ratio $\gamma$ are kept the same among the three variants. Table S\ref{tab:arkitecture} lists the detailed settings of these model architectures. 

% {\bf Parameter Complexity: } We provide a model parameter analysis in Appendix detailing and comparing the parameter complexity of  gMLP block and two implementations of PosMLP blocks based on GLRPE and GGQPE.
\vspace{-0.1cm}
\section{Experiments}
\label{sec:5}
In this section, we report the experimental results evaluated on ImageNet1K and COCO2017. We also conduct an ablation study to validate the impacts of hyperparameters and key components of our model. Moreover, we also give the visualized results to clearly observe the functions of the introduced positional encoding in spatial relation modeling. All the experiments are conducted on 4/8$\times$ 3090 GPUs.

\subsection{RPE in Vision MLP }
\label{sec:5.1}
We firstly give a comprehensive comparison for the proposed PoSGU modules. Both gMLP and our PosMLP are employed as backbone architectures. Table \ref{tab:1} lists their model complexity w.r.t token-mixing, extra FLOPs and top-1 accuracy by performing image classification on ImageNet1K$^{0.5}$. Note that the extra FLOPs means the additional computational cost of gating unit compared with base SGU.  

% Table \ref{tab:1} reports the performance of introducing aforementioned several variants of PoSGU into vanilla gMLP and our PosMLP. PoSGU/LRPE-M refers to the variant of Eq. (\ref{eq:9}) that with original projection matrix $\bm W$ and PoSGU/GLRPE is a group-wise variant of PoSGU/LRPE constructed follow Eq. (\ref{eq:13}).
\begin{table}[h]
  \vspace{-0.2cm}
    \centering
    \footnotesize
     \renewcommand{\multirowsetup}{\centering}
     \caption{Comparison of different RPE module variants with respect to mode complexity, extra FLOPs and top-1 accuracy using gMLP and our PosMLP as backbones. SGU refers to the spatial gating unit proposed by gMLP. PoSGU refers to our proposed positional spatial gating unit. PoSGU has four variants, i.e., LRPE-M, LRPE, and two corresponding group-wise modules GLRPE and GGQPE.}
     \vspace{-0.25cm}
     \begin{tabular}{|c|c|c|c|c|c|}
        \hline
        Model& \multicolumn{2}{c|}{Module} & \makecell[c]{Token-mixing \\complexity}&\makecell[c]{Extra FLOPs} &\makecell[c]{Top-1 acc.}  \\
        \Xhline{1.2pt}
        \multirow{4}{*}{gMLP}&\multicolumn{2}{c|}{SGU} &$\mathcal{O}(N^2)$& \XSolidBrush&72.14 \\\cline{2-6}
        &\multirow{4}{*}{PoSGU}&LRPE-M &$\mathcal{O}(N^2)$& $\mathcal{O}(N^2)$&73.96(+1.82) \\\cline{3-6}
        &&LRPE & $\mathcal{O}(N)$&$\mathcal{O}(N^2)$&72.44(+0.30)\\\cline{3-6}
        &&GLRPE & $\mathcal{O}(N)$&$\mathcal{O}(sN^2)$&\textbf{74.56}(+2.42)\\\cline{3-6}
        &&GGQPE &$\mathcal{O}(1)$ &$\mathcal{O}(sN^2)$&74.02(+1.88)\\
        \Xhline{1.2pt}
        \multirow{4}{*}{PosMLP}&\multicolumn{2}{c|}{SGU} &$\mathcal{O}(N^2)$& \XSolidBrush& 76.33 \\\cline{2-6}
        &\multirow{4}{*}{PoSGU}&LRPE-M &$\mathcal{O}(N^2)$&$\mathcal{O}(N^2)$& 76.95(+0.62) \\\cline{3-6}
        &&LRPE &$\mathcal{O}(N)$& $\mathcal{O}(N^2)$& 76.93(+0.60)\\\cline{3-6}
        &&GLRPE &$\mathcal{O}(N)$& $\mathcal{O}(sN^2)$&$\mathbf{77.41(+1.08)}$\\\cline{3-6}    
        &&GGQPE &$\mathcal{O}(1)$& $\mathcal{O}(sN^2)$&77.40(+1.07)\\

         \hline
        \end{tabular}\\
        \vspace{0.1cm}

    \label{tab:1}
    \vspace{-0.5cm}
\end{table}

\textbf{Observation and analysis.} As shown in this table, the proposed PoSGU modules consistently outperform SGU regardless of backbones. This provides a strong proof for the feasibility of the idea of manually parameterizing cross-token correlation. Specifically, LRPE, which is a simplified LRPE-M by removing the Token FC weights $\mathbf{W}$ in Eq. (9),  has almost the same top-1 accuracy (76.93 vs. 76.95 of LRPE-M for PosMLP) but significantly reduces the model complexity from $\mathcal{O}(N^2)$ to $\mathcal{O}(N)$. Moreover, the group-wise RPE versions GLRPE and GGQPE, which further emphasizes the multiple granularities of token contexts, achieve better performance than the non-group-wise counterparts. Finally, by considering the parameter efficiency, we choose GGQPE with only $\mathcal{O}(1)$ parameter complexity as the instantiation of our PoSGU. It is worth noting that the extra FLOPs of LRPE comes from the assignment operation while GQPE has an extra cost of Multiply-Add cumulation (MACs). However, as analyzed in Appendix, the extra FLOPs incurred by GGQPE is negligible compared with the model architecture block that has an $\mathcal{O}(rmdN^2)$ ($rmd>>s$, $s$ denotes the group number). Thus, for better illustration of direct parameterization of cross-token relation, we use GGQPE as our defalut setting for our PosMLP.

\vspace{-0.34cm}
% The introduing of LRPE to vanilla SGU gains significant promotion with negligible parameters. Though the removal of Token FC weights $\mathbf{W}$ restricts the performance to a extent (73.96->72.44 for gMLP, 76.95->76.93 for PosMLP), it still exhibits better than SGU while reducing the complexity. Finally the implementaion of both groudwise RPE significantly boost the performance. Considering the merit of constant-level complexity and better demonstrating the idea of manually parameterizing cross-token correlation, we set PoSGU/GGQPE as our default setting of our PoSGU as illustrated in Figure \ref{fig:1}. Notably, though GGQPE suffers from extra computational cost $\mathcal{O}(sN^2)$, where $s$ is the group number, it typically is negligible from block computational cost\footnote{$\gamma$ is the channel expansion ratio, $m$ represents windows number and $d$ is channel dimension} $\mathcal{O}(\gamma mdN^2)$

\begin{table}[th]
  \centering
  \small
  %  \footnotesize
      %\resizebox{\linewidth}{24mm}
      \setlength{\tabcolsep}{1.6mm}{
        \caption{Performance comparison of PosMLP variants with the state-of-the-arts such as CNNs, vision transformers and vision MLPs on ImageNet1K dataset.}
        \label{tab:imageclassification}
        \vspace{-0.25cm}
      \begin{tabular}{|c|ccc|c|} 
    \Xhline{1.2pt}
  \textbf{Method} &\textbf{Input Size} &\textbf{\#Param.} &\textbf{FLOPs} &\textbf{Top-1 Acc.} \\ \Xhline{1.2pt}
  \multicolumn{5}{|c|}{Tiny Models}\\\Xhline{1.2pt}
      RegNetY-4G \cite{radosavovic2020designing} &$224^2$ &21M &4.0G &80.0 \\
      Swin-T \cite{liu2021swin} &$224^2$ &29M &4.5G &81.3\\
      Nest-T \cite{zhang2021aggregating} &$224^2$ &17M &5.8G &\textbf{81.5}\\\Xhline{1pt}
     gMLP-S \cite{Pay} &$224^2$ &20M &4.5G &79.6 \\
     Hire-MLP-S \cite{guo2021hire}&$224^2$&33M&4.2G&81.8\\
      ViP-Small/7 \cite{hou2021vision}&$224^2$ &25M &6.9G &81.5 \\
   %   S2-MLPv2-Small/7 \cite{yu2021s} &$224^2$&25M&6.9G&82.0\\
   %   $\text{PosMLP}^-\text{-T}$(LRPE)  &$224^2$ &20M &5.0G &81.0 \\
    %  \textbf{$\text{PosMLP}^-\text{-T}$} &$224^2$ &19M &5.4G &81.9\\
     \textbf{PosMLP-T} &$224^2$ &21M &5.2G &\textbf{82.1}\\
     \textbf{PosMLP-T} &$384^2$ &21M &17.7G &\textbf{83.0}\\\Xhline{1.2pt}
     
  \multicolumn{5}{|c|}{Small Models}\\\Xhline{1.2pt}
      RegNetY-8G \cite{radosavovic2020designing} &$224^2$ &39M &8.0G &81.7 \\
     Swin-S \cite{liu2021swin} &$224^2$ &50M &8.7G &83.0\\
     Nest-S \cite{zhang2021aggregating} &$224^2$ &38M &10.4G &\textbf{83.3}\\\Xhline{1pt}
     Mixer-B/16 \cite{tolstikhin2021mlp}&$224^2$ &59M &11.7G &76.4\\
      S2-MLP-deep \cite{yu2021s}&$224^2$&51M&9.7G&80.7\\
      ViP-Medium/7 \cite{hou2021vision}&$224^2$ &55M &16.3G &82.7\\
      Hire-MLP-B \cite{guo2021hire}&$224^2$&58M&8.1G&\textbf{83.1}\\
      AS-MLP-S\cite{lian2021mlp}&$224^2$ &50M&8.5G&\textbf{83.1}\\
     \textbf{PosMLP-S} &$224^2$ &37M &8.7G &83.0\\\Xhline{1.2pt}
    
  \multicolumn{5}{|c|}{Base Models}\\\Xhline{1.2pt}
      RegNetY-16G \cite{radosavovic2020designing} &$224^2$ &84M &16.0G &82.9 \\ 
      Swin-B \cite{liu2021swin} &$224^2$ &88M &15.4G &83.3 \\
       Nest-B \cite{zhang2021aggregating} &$224^2$ &68M &17.9G &\textbf{83.8} \\\Xhline{1pt}
      gMLP-B \cite{Pay} &$224^2$ &73M &15.8G &81.6 \\ 
      ViP-Large/7 \cite{hou2021vision} &$224^2$ &88M&24.3G &83.2 \\
      Hire-MLP-L \cite{guo2021hire} &$224^2$&96M&13.5G&83.4\\
      \textbf{PosMLP-B} &$224^2$ &82M &18.6G &\textbf{83.6}\\\Xhline{1.2pt}  
  \end{tabular}}

  \vspace{-0.4cm}
\end{table}

\subsection{Image Classification}
\textbf{Dataset and implementation.} The image classification is benchmarked on the popular ImageNet1K \cite{deng2009imagenet} dataset. The training/validation partitioning follows the official protocol, where $\sim$1.2M images are for training and 50K images are for validation under 1K semantic categories. The top-1 accuracy (\%) is reported for performance comparison. In the \textbf{training}, we adopt the same data augmentations used in DeiT \cite{touvron2021training}, including RandAugment~\cite{cubuk2020randaugment}, Mixup~\cite{zhang2017mixup}, Cutmix~\cite{yun2019cutmix}, random erasing~\cite{zhong2020random} and stochastic depth~\cite{huang2016deep}. Exponential moving average (EMA)~\cite{polyak1992acceleration} is also used for training acceleration. We train the model for 300 epochs with batch-size 120 per GPU and the cosine learning rate schedule where the initial value is set to $1\times10^{-3}$ and minimal value is of $1\times10^{-5}$. The AdamW~\cite{loshchilov2017decoupled} optimizer with the momentum of 0.9 and weight decay of 0.067 is adopted. \textbf{Inference} is performed on a single $224\times 224$ center crop using the validation set.

\textbf{Results.} We compare our PosMLPs with current SOTAs on ImageNet1K, including the CNN-based ResNetY models, Transformer-based Nest and Swin models, and other vision MLPs, like gMLP and AS-MLP. Table \ref{tab:imageclassification} reports their performance comparison regarding parameters, computations (FLOPs) and top-1 accuracies. 

Amongst tiny models, our PosMLP-T achieves the 82.1\% top-1 accuracy, which is better than the results of CNN-based and Transformer-based, i.e., ResNetY-4G, Nest-T, and Swin-T. Benefiting from the hierarchical structure and replacing the vanilla Token FC layer with RPE based token-mixing layer, PosMLP-T requires fewer parameters and performed well on the tiny version. While under the small and base settings, our PosMLP still exhibits its strong competitiveness. Considering the model complexity, our PosMLP-S is more efficient than Swin-S (parameters: 37M vs 50M; FLOPs: 8.7G vs 8.7G). Compared to its mostly related gMLP, PosMLP variants consistently outperform the gMLP counterparts. For example, PosMLP-B outperforms gMLP-B by a margin of 2.0\%,
 which in a sense demonstrates the effectiveness of our implements in MLPs.

\subsection{Ablation Study}
\label{sec:4.2}
In this section, we separately study the impacts of the
model components, PoSGU variants, feature group number $s$ and window size. All the ablation study experiments are based on the tiny version of PosMLP and conducted on the $\text{ImageNet1K}^{0.5}$.

\begin{table}[htp]

  \centering
  \vspace{-0.2cm}
      \setlength{\tabcolsep}{0.8mm}{
        \caption{Comparison with different model components.}
        \label{tab:3}
        \vspace{-0.3cm}
        \begin{tabular}{|l|cccc|}\hline
      Model &  \#Param.&Flops& \makecell[c]{Top-1\\acc.}& \makecell[c]{Top-5\\acc.}\\\Xhline{1.2pt}
     gMLP (original) &19.4M&4.42G& 72.18 &90.52	    \\
     gMLP+GGQPE &18.2M&4.45G& 74.02&92.00 	    \\
     gMLP+\textit{ConvLHS}&21.8M&5.10G& 76.33& 93.07	    \\
     gMLP+\textit{ConvLHS}+GGQPE &20.9M&5.21G& \textbf{77.40} &	\textbf{93.58}     \\\hline
  \end{tabular}}
  \vspace{-0.05cm}
\end{table}

\textbf{Model components}
\label{sec:5.2}
 The key components in PosMLP can be generalized in twos: the convolution-linked hierarchical structure (ConvLHS) and the PoSGU components. ConvLHS composes the window-based architecture and convolutional operations (ConvPE + ConvPM) to realize efficient parameter sharing and local information modeling. As shown in Table \ref{tab:3}, we first perform the comparison between the original gMLP and the gMLP with the proposed ConvLHS. It can be found that gMLP+ConvLHS significantly improves gMLP by a large margin of $76.33-72.18=4.15$. Secondly, we directly replace the Token FC layers of gMLP with the proposed group-wise PosToken FC and observe 1.84 percentage points of performance improvement with fewer parameters and negligible extra FLOPs (we assign $s = 8$ for all layers), which echoes our statement that Token FC can be replaced with positional encoding for more efficient token mixing. Finally, the combination of GGQPE and ConvLHS results in the highest performance of 77.40.
 
%  we test positional encoding mechanisms GGQPE. These results demonstrate that the joint use of ConvLHS and positional encoding further introduce significant performance increasing (77.40 vs 76.33)

\begin{figure}

\begin{minipage}{1\linewidth}
  \small
      \centering
       \includegraphics[width=1\linewidth]{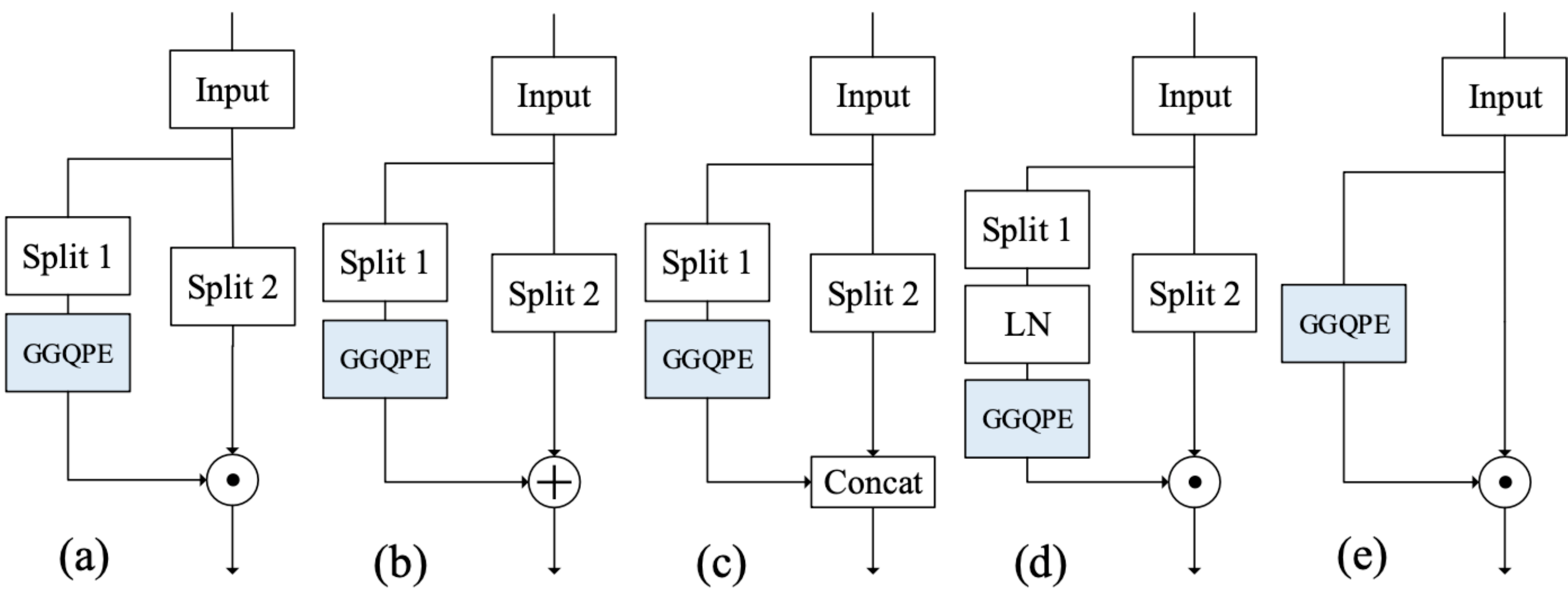}

  \end{minipage}
  \begin{minipage}[t]{1\linewidth}
      \centering
      \begin{tabular}{|c|c|c|c|c|c|}\hline
        PoSGU&(a)&(b)&(c)&(d)&(e)\\\Xhline{1.2pt}
        Top-1 acc.&$\mathbf{77.40^*}$& 76.97& 76.98&76.95&75.49\\\hline
      %   PoSGU& \makecell[c]{Top-1\\acc.}& \makecell[c]{Top-5\\acc.}\\\hline
      %   Addition & 76.97&93.46 \\ 
      %   Concatenation& 76.98&93.39 \\
      %    LayerNorm&76.95& 93.44\\
      %    NonSplit&75.49&92.70\\
      %    Standard & \textbf{77.40}& \textbf{93.58}\\
      % \hline
          \end{tabular}
          % \makeatletter\def\@captype{table}\makeatother\caption{4(a)}
        \end{minipage}
    % \makeatletter\def\@captype{table}\makeatother\caption{4(b)}
    \vspace{-0.14cm}
    \caption{PoSGU variants. (a) Standard; (b) Element-wise Addition; (c) Concat; (d) LayerNorm; (e) NonSplit. `*' denotes our defalut settings for PosMLP}
    \label{fig:5}
    \vspace{-0.4cm}   
\end{figure}
%  \begin{figure}[htb]
%   \centering
%    \includegraphics[width=1\linewidth]{figures/4units.pdf}
%     \caption{PoSGU variants. (a) Standard; (b) Element-wise Addition; (c) Concat; (d) LayerNorm; (e) NonSplit.}
%     \label{fig:5}
%     		\vspace{-0.5cm}
% \end{figure}

\textbf{PoSGU configurations}
The proposed PoSGU works in a gating manner as gMLP that splits the token embedding tensor into two parts. We study several potential configurations of PoSGU to verify the rationality of the design. Figure \ref{fig:5} illustrates the studied PoSGU configurations, including Element-wise Addition, Concat, LayerNorm and NonSplit variants, as well as their corresponding classification performances. Firstly, when replacing the gating operation with element-wise addition (subfigure-(b)) and channel concatenation (subfigure-(c)), we observe that performances degrade. This indicate the effectiveness of the gating operation. Secondly, when additionally adding a \textit{LayerNorm} before the GGQPE (subfigure-(d)), there is also a performance degradation (77.40$\rightarrow$76.95). The potential reason could be that the \textit{Softmax} used by GGQPE (see Eq. (\ref{eq:ggqpe})) contribute to normalized token-mixed information, and extra \textit{LayerNorm} will break the feature homogeneity for two splits. Finally, we empirically omit the channel splitting operation (subfigure-(e)) and find the same trend observed by gMLP that channel splitting is more effective (75.49 vs. 77.40).
\begin{table}[htp]
  \vspace{-0.3cm}
  \centering
  \caption{Comparison with different group numbers $s$.}
  \vspace{-0.3cm}
  \begin{tabular}{|c|cc|}\hline
      % $s$&$1/1/1/1$&$4/4/4/4$ &$4/8/16/32$ &$16/32/64/128$ &$32/32/32/32$ &$8/16/32/64$\\

       $s$ each stage & \makecell[c]{Top-1 acc.}& \makecell[c]{Top-5 acc.}\\\Xhline{1.2pt}
       $(1,1,1,1)$ &76.70& 93.02        \\
       $(4,4,4,4)$ & 76.87 &93.40 	    \\
        $(4,8,16,32)$ & 77.10 	&93.24    \\
      %  $(16,32,64,128)$ & 77.15&93.54 \\
       $(32,32,32,32)$ & 77.02 &93.14	    \\
        $(8,16,32,64)^*$&   $\mathbf{77.40}^*$&  {$\mathbf{93.58}^*$}  \\\hline
       
    \end{tabular}
    % \makeatletter\def\@captype{table}\makeatother\caption{4(b)}
\label{tab:4}
\vspace{-0.35cm}
\end{table}
  % \vspace{0.6cm}

  \textbf{Group number $s$.}
We compare different settings of group number $s$, which is introduced for group operation in the proposed PosToken FC module (Here we inplemented on GGQPE design), in Table \ref{tab:4}. We observe that using group operation (76.70$\rightarrow$76.87$\rightarrow$77.02) and hierarchically increasing it (76.87$\rightarrow$77.10$\rightarrow$77.40) in each stage can significantly improve the performance. By jointly considering the channel dimensions in different stages and the performance, we finally set $s=8\times StageID$ for each stage, leading to $(8, 16, 32, 64)$ for stages 1/2/3/4.

\begin{table}[htp]
  \centering
      \setlength{\tabcolsep}{1.2mm}{
        \caption{Comparison using different window sizes $(k, k)$.}
        \label{tab:5}
        \vspace{-0.3cm}
        \begin{tabular}{|c|cccc|}\hline
     \makecell[c]{Win size (
     $1_{th}$ stage)} &\#param.&FLOPs &   \makecell[c]{Top-1 acc.}& \makecell[c]{Top-5 acc.}\\ \Xhline{1.2pt}
      $28\times28$&20.8M&5.95G& \textbf{77.81} &\textbf{93.91}	\\
      $14\times14^*$&20.8M&5.21G& $77.40^*$ &$93.58^*$	\\
      $7\times7$&20.8M&5.01G& 76.98 &93.45	\\\hline
  \end{tabular}
}
  \vspace{-0.3cm}
\end{table}

\textbf{window size.}
The window size decides the number of tokens $N$ in a PosMLP block. We compare three settings in the first stage, i.e., $7\times 7$, $14\times 14$ and $28\times 28$. From Table \ref{tab:5}, it can be found that larger window size leads to monotonic performance increasing and also dramatically raises the FLOPs. To consider the trade-off between performance and efficiency, we set the patch window size as $14\times 14$ for all model variants.

% \begin{minipage}{1\linewidth}
%   \begin{minipage}[t]{0.45\linewidth}
%     \centering
% 		\begin{tabular}{|cc|cc|}\hline
%       $\mathbf{\Sigma}$&$\mathbf{\Delta}$ &\makecell[c]{Top-1 acc.}&\makecell[c]{Top-5 acc.}\\\hline
%       $\alpha \mathbf{I}$&\checkmark&76.49&93.20\\
%       $\bm \Gamma$&\checkmark& \textbf{77.61}&\textbf{93.90}\\
%       $\bm \Gamma\bm \Gamma^\top$&\checkmark&$77.40^{*}$&$93.58^*$\\
%       $\bm \Gamma\bm \Gamma^\top$&\XSolidBrush&77.20&93.40\\
%       \hline
%           \end{tabular}
%   \end{minipage}  
%   \begin{minipage}[t]{0.55\linewidth}
%     \centering
%     \begin{tabular}{|cc|c|}\hline

%        Bias $\bm b$ & APE & \makecell[c]{Top-1\\acc.}\\\hline
%     \XSolidBrush&\XSolidBrush& 76.8\\
%            \XSolidBrush&\checkmark&77.3\\
%            \checkmark&\XSolidBrush&$\mathbf{77.4}^*$\\
%            \checkmark&\checkmark&77.3\\\hline

%       \end{tabular}
%   \end{minipage}

%   \makeatletter\def\@captype{table}\makeatother\caption{Left (a) shows ablation study of changing the property of covariance matrices $\bm\Sigma$ and whether freeze shift attention center $\bm\Delta$ or not. Right (b) ablates on the relationship between bias term in Eq.(\ref{eq:13}) and absolute positional encoding (APE) in PosMLP-T. '*' denotes our defalut settings for PosMLP. }
%   \label{tab:6}
% \end{minipage}

\begin{figure}[tp]
  \centering
\begin{subfigure}{0.32\linewidth}
\centering
\includegraphics[width=1\linewidth]{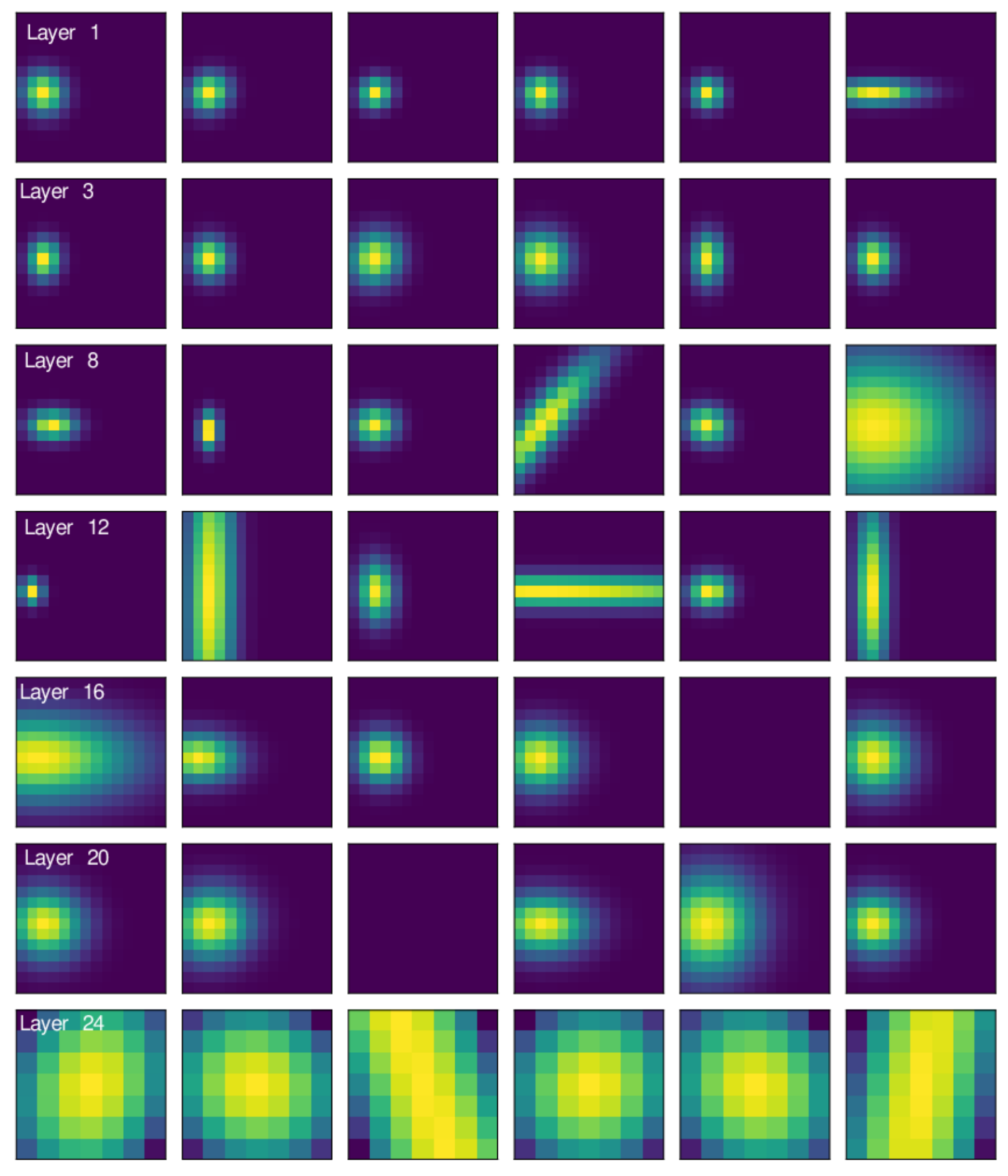}
  \caption{$\mathbf{\Sigma} = \mathbf{\Gamma}\mathbf{\Gamma}^\top$} 
  \label{fig:vam_gg}
  \end{subfigure}
\begin{subfigure}{0.32\linewidth}
\centering
\includegraphics[width=1\linewidth]{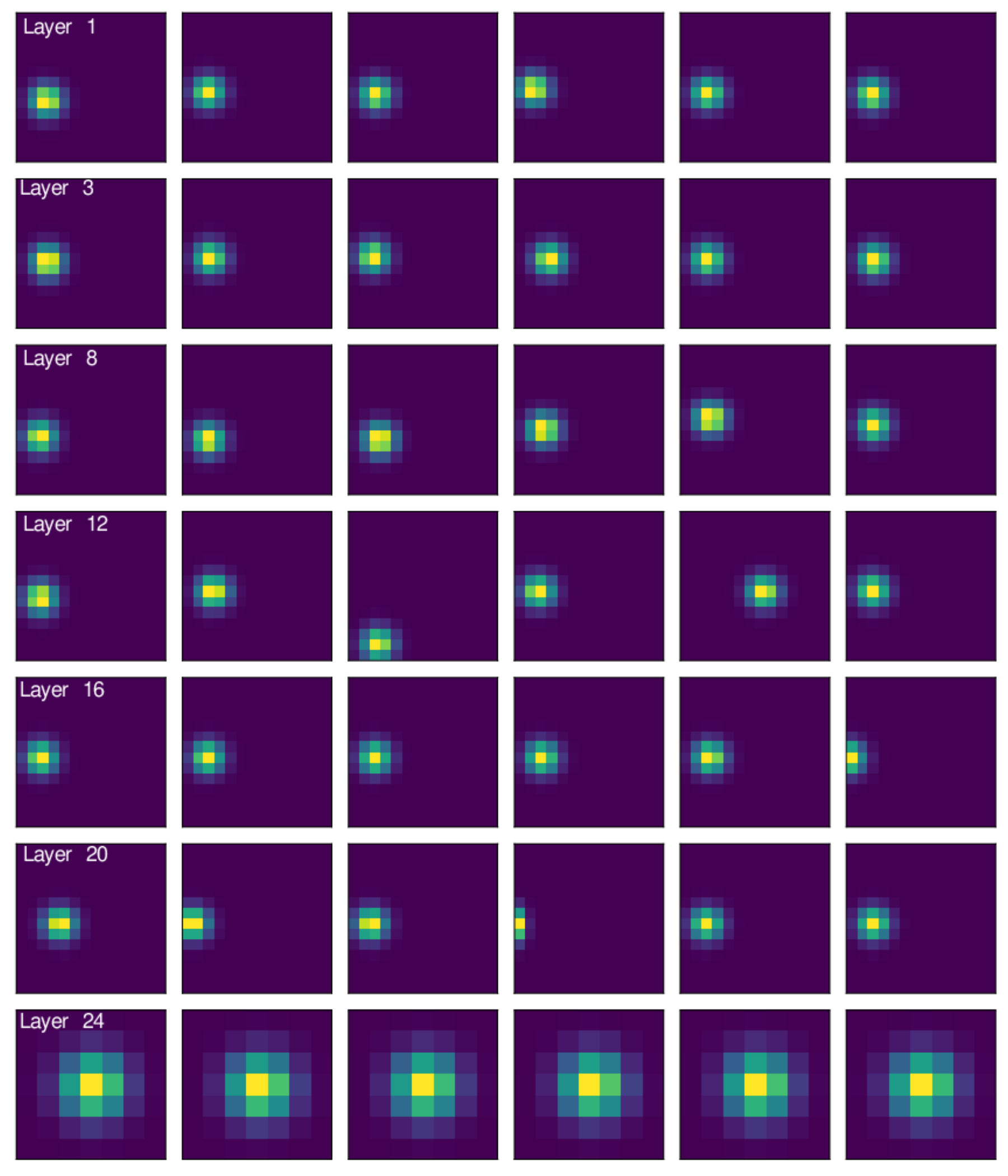}
  \caption{$\mathbf{\Sigma}=\alpha \mathbf{I}$} 
  \label{fig:vam_ai}
  \end{subfigure}
\begin{subfigure}{0.32\linewidth}
\centering
\includegraphics[width=1\linewidth]{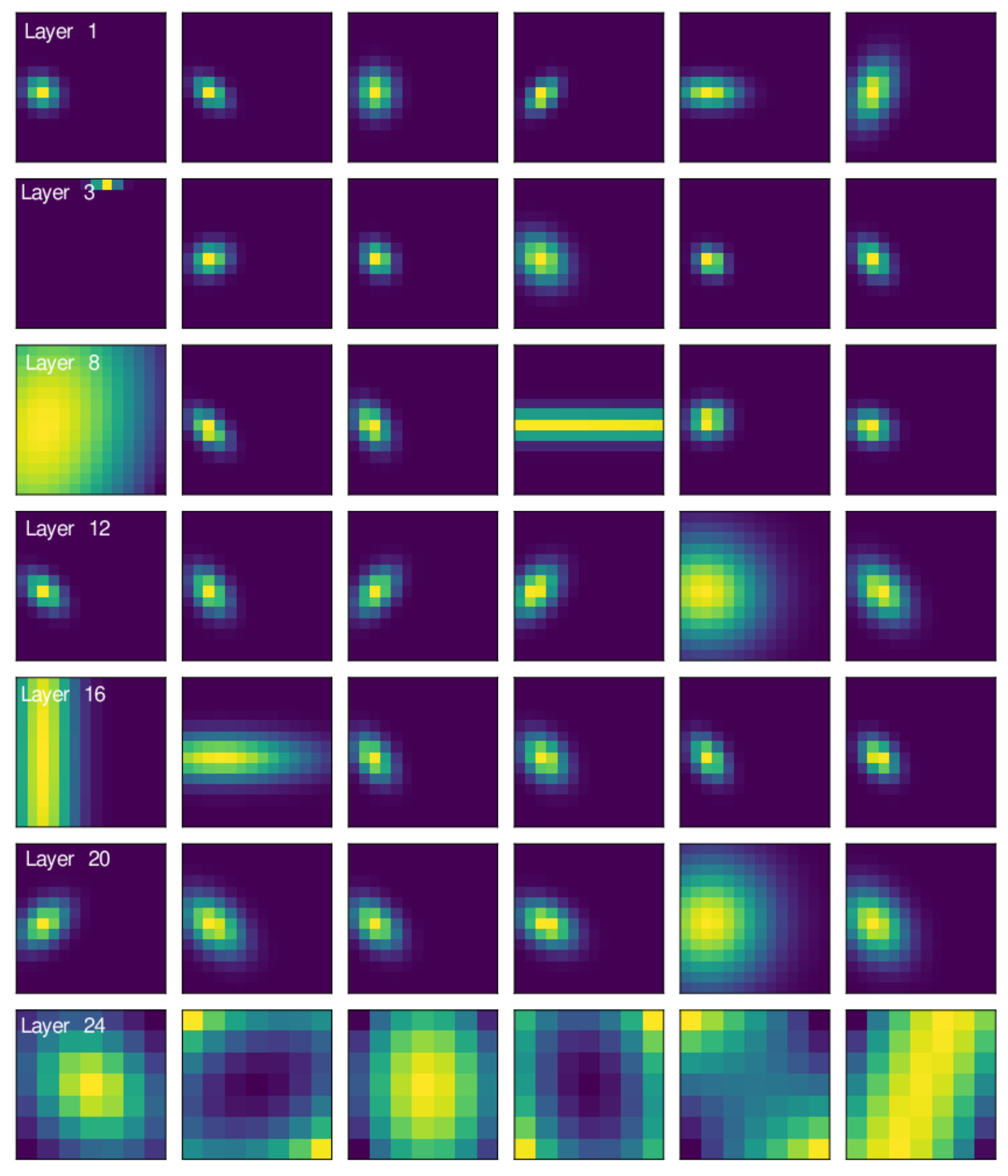}
  \caption{$\mathbf{\Sigma} = \mathbf{\Gamma}$} 
  \label{fig:vam_g}
\end{subfigure}
\vspace{-0.23cm}
  \caption{Visualization of attention logits for a given query token with different form of covariance matrix $\mathbf{\Sigma}$. Row represents different layers and column represents the selected groups in GGQPE. Results is based on PosMLP-T}
      \vspace{-0.4cm}
  \label{fig:vam}
\end{figure}

\begin{table}[htp]
  \centering
  \caption{ Ablation study of changing the property of covariance matrices $\bm \tilde{\Sigma}$ and whether freeze $\bm\Delta$ or not. }
  \vspace{-0.3cm}
  \begin{tabular}{|c|c|c|c|c|}\hline
    $\mathbf{\Sigma}$ form&$\alpha \mathbf{I}$&$\bm \Gamma$& $\bm \Gamma\bm \Gamma^\top$&$\bm \Gamma\bm \Gamma^\top$\\\hline
    $\mathbf{\Delta}$ freezed&\checkmark&\checkmark&\checkmark&\XSolidBrush\\\Xhline{1.2pt}
    Top-1 acc. & 76.49&\textbf{77.61}&$77.40^{*}$&77.20\\\hline
    % $\mathbf{\Sigma}$&$\mathbf{\Delta}$ &\makecell[c]{Top-1 acc.}&\makecell[c]{Top-5 acc.}\\\hline
    % $\alpha \mathbf{I}$&\checkmark&76.49&93.20\\
    % $\bm \Gamma$&\checkmark& \textbf{77.61}&\textbf{93.90}\\
    % $\bm \Gamma\bm \Gamma^\top$&\checkmark&$77.40^{*}$&$93.58^*$\\
    % $\bm \Gamma\bm \Gamma^\top$&\XSolidBrush&77.20&93.40\\
        \end{tabular}
        \label{tab:6}
        \vspace{-0.3cm}
      \end{table}
\textbf{$\mathbf{\Sigma}$ and $\mathbf{\Delta}$ terms in GGQPE.} 
\label{sec:4.2.4}
Recall that the positional weight $\mathbf{W}^{\textmd{gqpe}}$ in Eq. (\ref{eq:ggqpe}) is determined by two terms: covariance matrix $\mathbf{\Sigma}$ and shift attention center $\mathbf{\Delta}$. Firstly, we investigate three different implementations for the covariance matrix $\mathbf{\Sigma}$, i.e., $\mathbf{\Sigma}=\alpha \mathbf{I}$ where $\mathbf{I}$ denotes the identity matrix, $\mathbf{\Sigma}=\mathbf{\Gamma}$ and the standard $\mathbf{\Sigma}=\mathbf{\Gamma}\mathbf{\Gamma}^\top$. Specifically, $\mathbf{\Sigma}=\alpha \mathbf{I}$, which is known as Quadratic Encoding \cite{Onthe}, only forms isotropic Gaussian distributions. By contrast, both $\mathbf{\Sigma}=\mathbf{\Gamma}$ and $\mathbf{\Sigma}=\mathbf{\Gamma}\mathbf{\Gamma}^\top$ can learn non-isotropic Gaussian distributions over tokens. The positive semi-definite matrix $\mathbf{\Gamma}\mathbf{\Gamma}^\top$ makes the  attention weights maximized at the learned shift attention center $\mathbf{\Delta}$, while under $\mathbf{\Sigma}=\mathbf{\Gamma}$ this will not be guaranteed any more. Among the three settings, as shown in Table \ref{tab:6}, $\mathbf{\Sigma}=\mathbf{\Gamma}$ and $\mathbf{\Sigma}=\mathbf{\Gamma}\mathbf{\Gamma}^\top$ can significantly outperform $\mathbf{\Sigma}=\alpha \mathbf{I}$, which demonstrates the superior of non-isotropic Gaussian distribution in attention learning. The three attention maps also visually give cues that $\mathbf{\Sigma}=\mathbf{\Gamma}$ and $\mathbf{\Sigma}=\mathbf{\Gamma}\mathbf{\Gamma}^\top$ successfully learn flexible local\&non-local patterns, whereas the $\mathbf{\Sigma}=\alpha \mathbf{I}$ only learn a relative local information, as shown in Figure \ref{fig:vam}. For the deep layers in $\mathbf{\Sigma}=\mathbf{\Gamma}\mathbf{\Gamma}^\top$ same groups learn a global average pool attention map, potently supporting our original intention of preserving the non-localilty of MLPs. Secondly, since a learnable $\mathbf{\Delta}$ can lead to floating shift attention center, we thus ablate on what if freezing $\mathbf{\Delta}=(0,0)$ which meats the shifted attention center is fixed to the anchor pixel itself. As shown in the table, the limited displacement flexibility causes a certain degree of but not fatal degradation.
\par

\vspace{-0.2cm}
\begin{table}[hp]

  \centering
  \caption{Ablation study on the relationship between bias term in Eq.(\ref{eq:13}) and absolute positional encoding (APE).}
  \vspace{-0.3cm}
  \begin{tabular}{|c|c|c|c|c|}\hline
    Bias $\bm b$ &\XSolidBrush&\XSolidBrush&\checkmark&\checkmark\\\hline
    APE & \XSolidBrush&\checkmark&\XSolidBrush&\checkmark\\\Xhline{1.2pt}
    Top-1 acc.& 76.8&77.3&$\mathbf{77.4}^*$&77.3\\\hline
%     Bias $\bm b$ & APE & \makecell[c]{Top-1\\acc.}\\\hline
%  \XSolidBrush&\XSolidBrush& 76.8\\
%         \XSolidBrush&\checkmark&77.3\\
%         \checkmark&\XSolidBrush&$\mathbf{77.4}^*$\\
%         \checkmark&\checkmark&77.3\\\hline
   \end{tabular}
   \vspace{-0.5cm}

  \end{table}

\begin{table*}[tp]

  \centering
  \caption{Performance comparison with state-of-the-arts on object detection using COCO2017 dataset.}
  \vspace{-0.2cm}
      \setlength{\tabcolsep}{1mm}{
  \begin{tabular}{|l|c|ccc|ccc|c|ccc|ccc|}
  \hline\multirow{2}{*}{Backbone} & \multirow{2}{*}{\#Param.}&\multicolumn{6}{c|}{Mask R-CNN $1\times$}&\multirow{2}{*}{\#Param.}&\multicolumn{6}{c|}{RetinaNet $1\times$}\\\cline{3-8}\cline{10-15}
  &&AP&$\text{AP}_{50}$&$\text{AP}_{75}$&$\text{AP}^\text{m}$&$\text{AP}^\text{m}_{50}$&$\text{AP}^\text{m}_{75}$&&AP&$\text{AP}_{50}$&$\text{AP}_{75}$&$\text{AP}_{S}$&$\text{AP}_{M}$&$\text{AP}_{L}$\\\hline
  
  ResNet50\cite{he2016deep}&44.2M&38.0&58.6&41.4&34.4&55.1&36.7&37.7M&36.3&55.3&38.6&19.3&40.0&48.8\\
    PVT-Small\cite{wang2021pyramid}&44.1M&40.4&62.9&43.8&37.8&60.1&40.3&34.2M&40.4&61.3&43.0&25.0&42.9&55.7\\
    CycleMLP-B2\cite{chen2021cyclemlp}&46.5M&41.7&63.6&45.8&38.2&60.4&41.0&36.5M&40.9&61.8&43.4&23.4&44.7&53.4\\\hline
  % 	PosMLP-T&40.5M&40.2&62.4&43.5&37.2&59.5&39.3&31.1M&39.2&59.6&41.9&23.7&42.4&52.6\\
    % PosMLP-T&40.5M&40.5&63.2&43.9&37.5&60.1&39.7&31.1M&39.9&60.4&42.5&24.3&43.3&53.0\\
    PosMLP-T(ours)&40.5M&41.6&64.1&45.6&38.4&61.1&41.0&31.1M&41.9&63.2&44.7&25.1&45.7&55.6\\\hline

    ResNet101\cite{he2016deep}&63.2M&40.4&61.1&44.2&36.4&57.7&38.8&56.7M&38.5&57.8&41.2&21.4&42.6&51.1\\
    PVT-Medium\cite{wang2021pyramid}&63.9M&42.0&64.4&45.6&39.0&61.6&42.1&53.9M&41.9&63.1&44.3&25.0&44.9&57.6\\
    CycleMLP-B3\cite{chen2021cyclemlp}&58.0M&43.4&65.0&47.7&39.5&62.0&42.4&48.1M&42.5&63.2&45.3&25.2&45.5&56.2\\\hline
  % 	PosMLP-T&40.5M&40.2&62.4&43.5&37.2&59.5&39.3&31.1M&39.2&59.6&41.9&23.7&42.4&52.6\\
    % PosMLP-T&40.5M&40.5&63.2&43.9&37.5&60.1&39.7&31.1M&39.9&60.4&42.5&24.3&43.3&53.0\\
    PosMLP-S(ours)&56.1M&43.2&65.5&47.4&39.4&62.5&42.1&47.3M&42.4&63.6&45.1&26.5&45.7&56.3\\\hline
  % 	ResNet101\cite{he2016deep}&63.2&40.4&61.1&44.2&36.4&57.7&38.8&56.7&38.5&57.8&41.2&21.4&42.6&51.1\\
  % 	ResNeXt101-32x4d\cite{xie2017aggregated}&62.8&41.9&62.5&45.9&37.5&59.4 &40.2&56.4&39.9&59.6&42.7&22.3&44.2&52.5\\
  % 	PVT-Medium\cite{wang2021pyramid}&63.9&42.0&64.4&45.6&39.0&61.6&42.1&53.9&41.9&63.1&44.3&25.0&44.9&57.6\\
  % 	PosMLP(ours)&56.2&40.7&62.6&44.1&37.3&59.8&39.5&47.4&&&&&&\\

  \end{tabular}}
  % \vspace{-0.2cm}
  \label{tab:7}
  \vspace{-0.31cm}
\end{table*}
\subsection{Bias May Reveal the Absolute Positional Information}
The proposed GGQPE explicitly learns the relative positional relation between a pair of tokens/pixels. 
Here we rewrite Eq. (\ref{eq:11}) with the presence of the bias term $b\in\mathbb{R}^{N\times1}$. The bias $b$ is an feature agnostic term and can assign an offset value to each pixel/token at the $\sqrt{N}\times \sqrt{N}$ feature map.

  \vspace{-0.3cm}
\begin{equation}  
  \vspace{-0.05cm}
  \mathbf{Z}^{\textmd{gqpe}^s} = (\mathbf{W}^{\textmd{gqpe}^s}\mathbf{X}^{1}_{s}+\bm b)\odot \mathbf{X}^{2}_{s}.
      \label{eq:13}
\end{equation}

\begin{figure}[tp]
  \centering
\begin{subfigure}{0.45\linewidth}
\centering
\fbox{\includegraphics[width=1\linewidth]{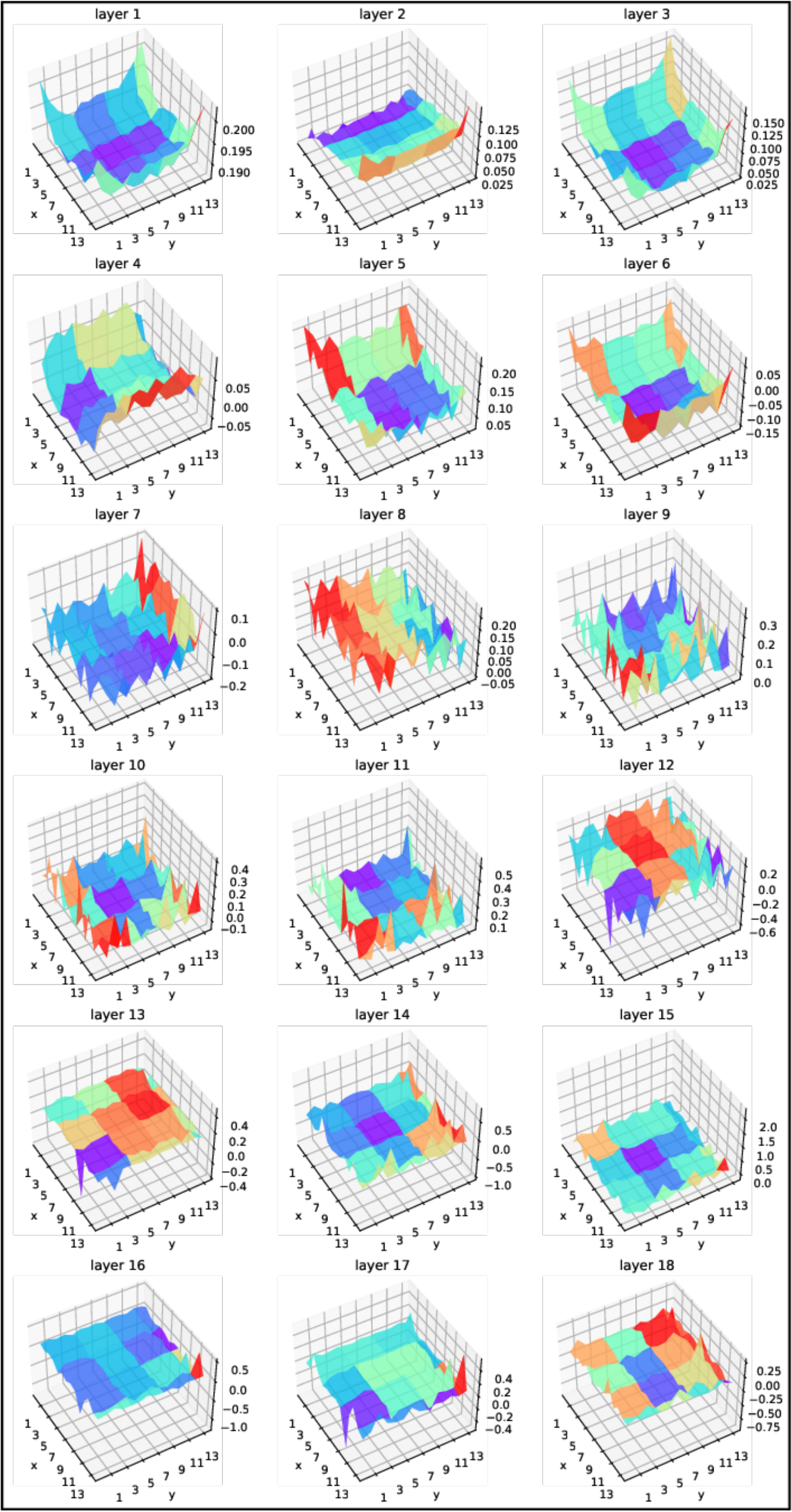}}
  \caption{gMLP/SGU} 
  \end{subfigure}
  \hspace{0.2cm}
\begin{subfigure}{0.45\linewidth}
\centering
\fbox{\includegraphics[width=1\linewidth]{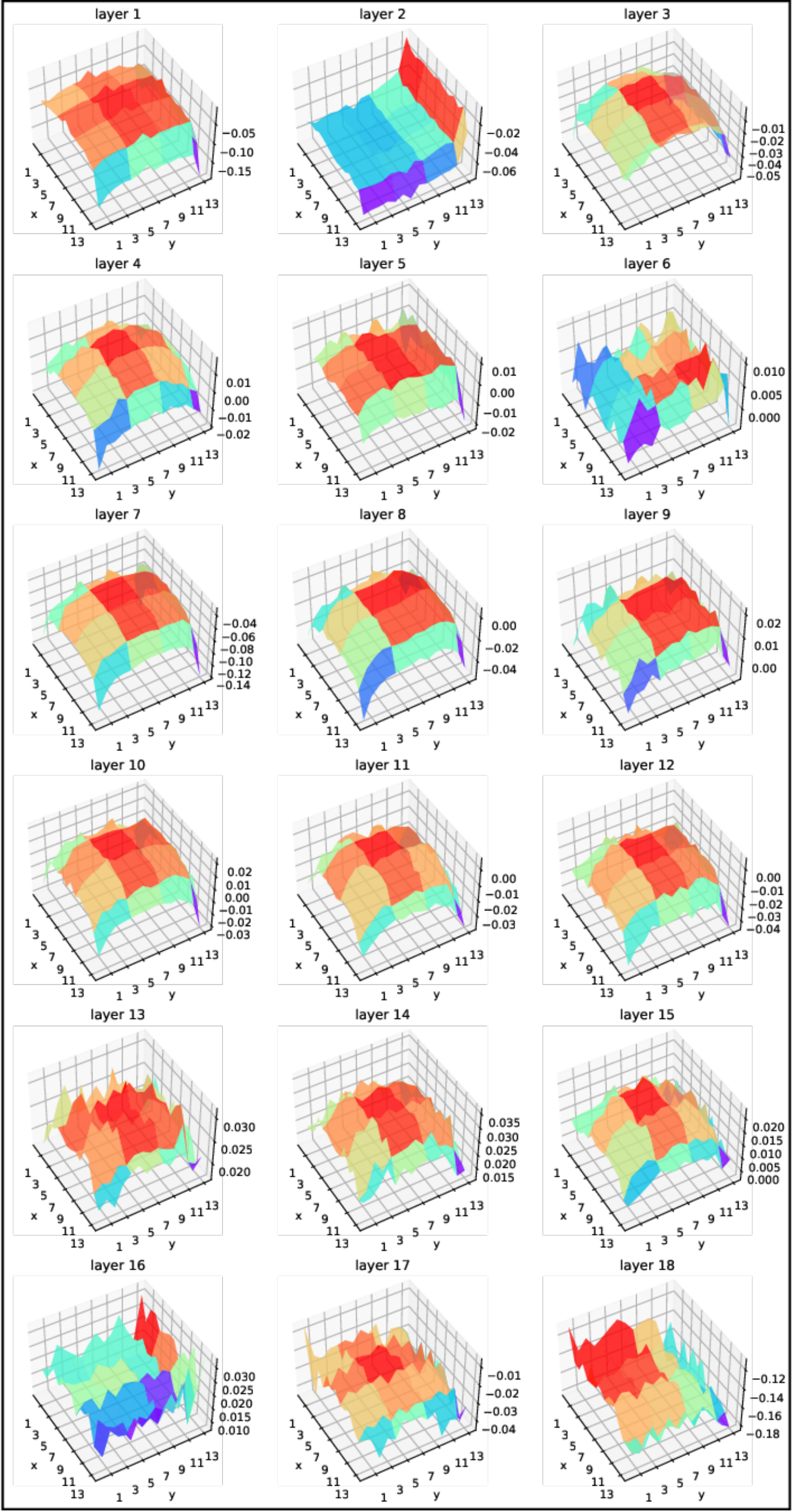}}
  \caption{PosMLP/PoSGU} 
  \end{subfigure}
  \vspace{-0.2cm}
  \caption{Visualization of logits map of $\sqrt{N}\times \sqrt{N}$ bias map. (a) comes from first 18 Token FC layers (blocks) pretrained from gMLP-S and (b) uses the 18 layers of 3$th$ stages in our pretrained PosMLP-T.}
      \vspace{-0.5cm}
  \label{fig:6}
\end{figure}

So far, we still have the question that whether GGQPE can capture absolute positional information of pixels and how important the absolute position is to PosMLP. To answer it, we purposefully add the absolute position embedding to the tokens following ViT \cite{dosovitskiy2020image}, that is, element-wisely adding a learnable parameter to each element of the token embedding before the first stage, resulting in a $\sqrt{N}\times\sqrt{N}\times C$ learnable matrix. Meanwhile, we also remove the bias term $\bm b$ in Eq. (\ref{eq:13}) so that the GGQPE focuses entirely on the relative positional encoding. Table \ref{tab:7} shows the performance changes of PosMLP w/ and w/o $\bm b$ and absolute position encoding (APE). It can be found that (1) PosMLP w/o $\bm b$ underperforms PosMLP w/ APE by 0.5 percent (76.8 vs 77.3), (2) PosMLP w/ $\bm b$ and PosMLP w/ APE have almost the same results (77.4 and 77.3), and (3) PosMLP w/ both $\bm b$ and APE does not obtain any performance gain. Based on these observations, we speculate that (1) the absolute position encoding is indeed essential for our PosMLP when modeling images and (2) the bias term $\bm b$ in GGQPE can achieve similar function with APE which explains why our PosMLP does not need any extra absolute positions. As shown in Figure \ref{fig:6}, we plot the logits intensity of bias $\bm b$ of SGU and PoSGU, respectively. We observe that the center area of an image generally has high bias values in the most stages of PoSGU, which is reasonable as the most of the informative objects locate at the center area in the ImageNet dataset.
   
\subsection{Object Detection} % and Instance Segmentation

\textbf{Dataset and setting.} We further examine our model on object detection using COCO2017 dataset~\cite{lin2014microsoft}. COCO2017  has 118k training images and 5k validation images. The implementation is based on the \textit{mmdetection}~\cite{chen2019mmdetection} package, where two classic object detection frameworks, \textit{i.e}. RetinaNet~\cite{lin2017focal} and Mask R-CNN~\cite{he2017mask}, are used. Here, the training and validation settings follow the basic protocol of Mask R-CNN and RetinaNet: 1$\times$ training scheduler (i.e., 12 epochs), batch-size 2 per GPU, resizing an image to the shorter side 800 and the longer side at most 1333, AdamW \cite{loshchilov2017decoupled} optimizer with the initial learning rate of $1\times 10^{-4}$ and weight decay of $0.067$.

\textbf{Implementation.} We adopts windows shifting operation in PosMLP-T following Swin Transformer \cite{liu2021swin} since for the large resolution dataset the non-overlap window partition will cause performance degeneration. Generally, the windows shifting operation can smooth partitioning traces and enhance windows interactions demonstrated by Swin Transformer.

\textbf{Results.} Table \ref{tab:7} shows the object detection results as well as the performance comparison with PVT \cite{wang2021pyramid}, ResNet \cite{he2016deep} and CycleMLP \cite{chen2021cyclemlp} that have similar computational complexity with our PosMLP. All PosMLP variants achieve consistent better performances than the standard convolutional networks. Secondly, PosMLP outperforms Transformer-based PVT-small on most metrics and obtain comparable results to CycleMLP but requires less parameters (e.g., 40.5M vs 46.5M, 31.1M vs 36.5M). 

\vspace{-0.20cm}
\section{Conclusion}

In this work, we have presented a new gating unit PoSGU and used it as the key building block to develop a new  vision MLP architecture referred to as the PosMLP. It has the advantage of reduced  parameter complexity  but without sacrificed model expressive power. We have adopted two RPE mechanisms and proposed the group-wise extension to boost their performance, 
%From TT: Check this.
and these improve the weak locality and single granular non-locality in the original vision MLP design. We have conducted thorough experiments to evaluate the proposed approach,
%From TT: Add some result highlight?
where the PosMLP exhibits high parameter and sample efficiency, e.g., Table \ref{tab:1} and Figure \ref{fig:1} (f). 
In the current version, we have demonstrated the merit of direct parameterization of cross-token relations for vision MLP. However, for an even larger scale pre-train dataset, a more flexible combination between RPE and TokenFC should be carefully designed, i.e., the trade-off between inductive bias and capability. We also hope this work will inspire further theoretical study of positional encoding in vision MLPs and could have a mature application as in vision Transformers.
\vspace{-0.1cm}
\section{ACKNOWLEDGEMENT}

The work was supported in part by the National Key Research and Development Program of China (2020YFB1406703), and by the National Natural Science Foundation of China (U21B2026).

\bibliographystyle{ACM-Reference-Format}
\balance 
\bibliography{sample-base}

\clearpage
\appendix

\section*{Appendix}

\begin{figure}[htp]
  \centering
  \includegraphics[width=1\linewidth]{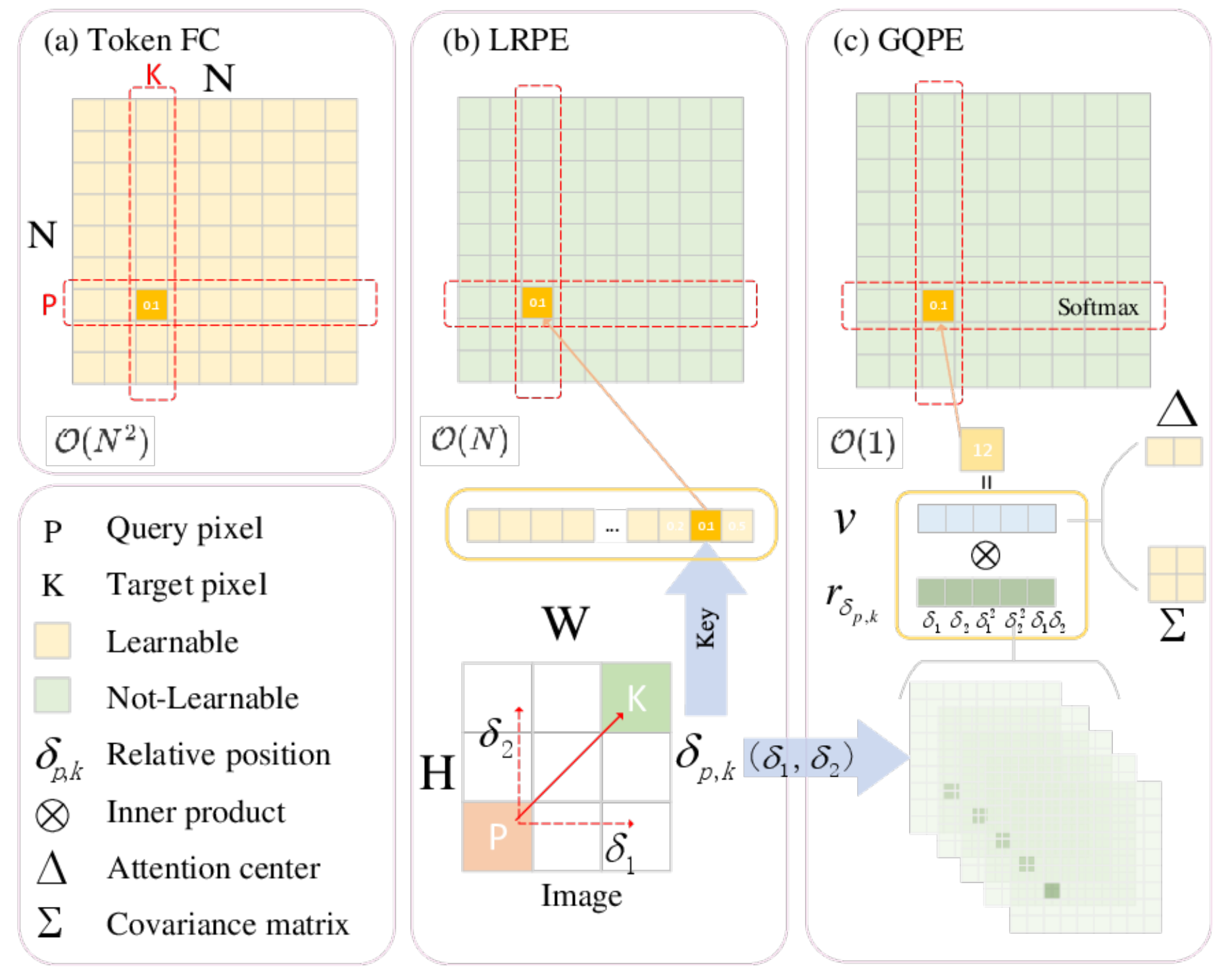}
  \caption{The demonstration of three token mixing method mentioned in this work. (a) Token FC needs to pairwisely compare the $N$ tokens; (b) The LRPE builds a learnable relation dictionary; (c) The GQPE learns the relation matrix with a constant number of parameters that is independent of $N$.}
  \label{fig:7}
\end{figure}

\subsection{Bias and APE in Object Detection}
We are curious the effect of bias term and additional absolute positional encoding (APE) in the Object Detection task (under the Mash-RCNN 1x setting). This result is presented in Table \ref{ape:ob}, and it is generally consistent with the observation presented in Swim Transformer \cite{liu2021swin}, that APE is effective in image classification while it will not promote the performance in objection detection. Instead in their implementation, the extra APE will slightly degrade the performance in object detection because the APE does harm to the translation invariance . According to this phenomenon, one possible explanation is that, in comparison with the non-overlap downsample strategy adopted in Swim, we are using a convolutional downsample strategy that inherits the nature of translation invariance and will weaken the inductive bias at the end of each stage. Besides, APE is playing an important role in DETR \cite{carion2020end} in contrast to Swim and our observation, we argue the possible explanation might be the use of multi-stage FPN and convolution-based RPN rather than the end-to-end transformer-based structure in DETR.
\begin{table}[hp]
\caption{ Ablation study on the bias term and absolute positional encoding (APE) in objection detection.}
\begin{tabular}{|c|c|c|c|c|}\hline
  Bias $\bm b$ &\XSolidBrush&\XSolidBrush&\checkmark&\checkmark\\\hline
  APE & \XSolidBrush&\checkmark&\XSolidBrush&\checkmark\\\Xhline{1.2pt}
  $AP^{box}$& 41.6&41.6&41.6&$\mathbf{41.7}$\\\hline
  $AP^{mask}$& 38.3&38.4&$\mathbf{38.5}$&38.4\\\hline
 \end{tabular}
 \vspace{-0.4cm}
\label{ape:ob}
\end{table}
\subsection{Convolutional Downsample Module}
We add the ablation study at different downsampling strategy, i.e., Patch Embedding (PE) module and Patch Mearging (PM) moudle. We here demonstrates the effectiveness of using overlapping downsampling (convolutional mapping) comparing with non-overlapping downsampling (Linear mapping). Using non-overlap PM tends to introduce more redundant parameters, and substitute it with single depthwise convolution operation will siginificantly decrease both the parameter and computation complexity. The overlap PE module will slightly increase computation cost, but compared with base model it improves the accuracy of 0.75$\%$ 

\begin{table}[htp]

  \centering
  \vspace{-0.15cm}
      \setlength{\tabcolsep}{0.8mm}{
        \caption{Performance comparison with different downsampling strategy for gMLP (with SGU).}
        \label{tab:14444}
        \vspace{-0.3cm}
        \begin{tabular}{|l|c|c|ccc|}\hline
      Architecture & PE & PM& \#Param.&Flops& \makecell[c]{Top-1\\acc.}\\\Xhline{1.2pt}
     Original&None&None &19.4M&4.42G& 72.18     \\\hline
     \multirow{4}{*}{Hierarchical} &Linear&Linear&23.3M&5.10G& 74.68 \\
     &Conv&Linear&23.3M&5.27G& 75.43\\
     &Linear&Conv&21.8M&4.93G & 75.81\\
     &Conv&Conv&21.8M&5.10G& $\mathbf{76.33}^*$\\\hline
  \end{tabular}}
  \vspace{-0.25cm}
\end{table}


\begin{figure}[thp]
  \centering
  \begin{subfigure}{0.47\linewidth} \includegraphics[width=1\linewidth]{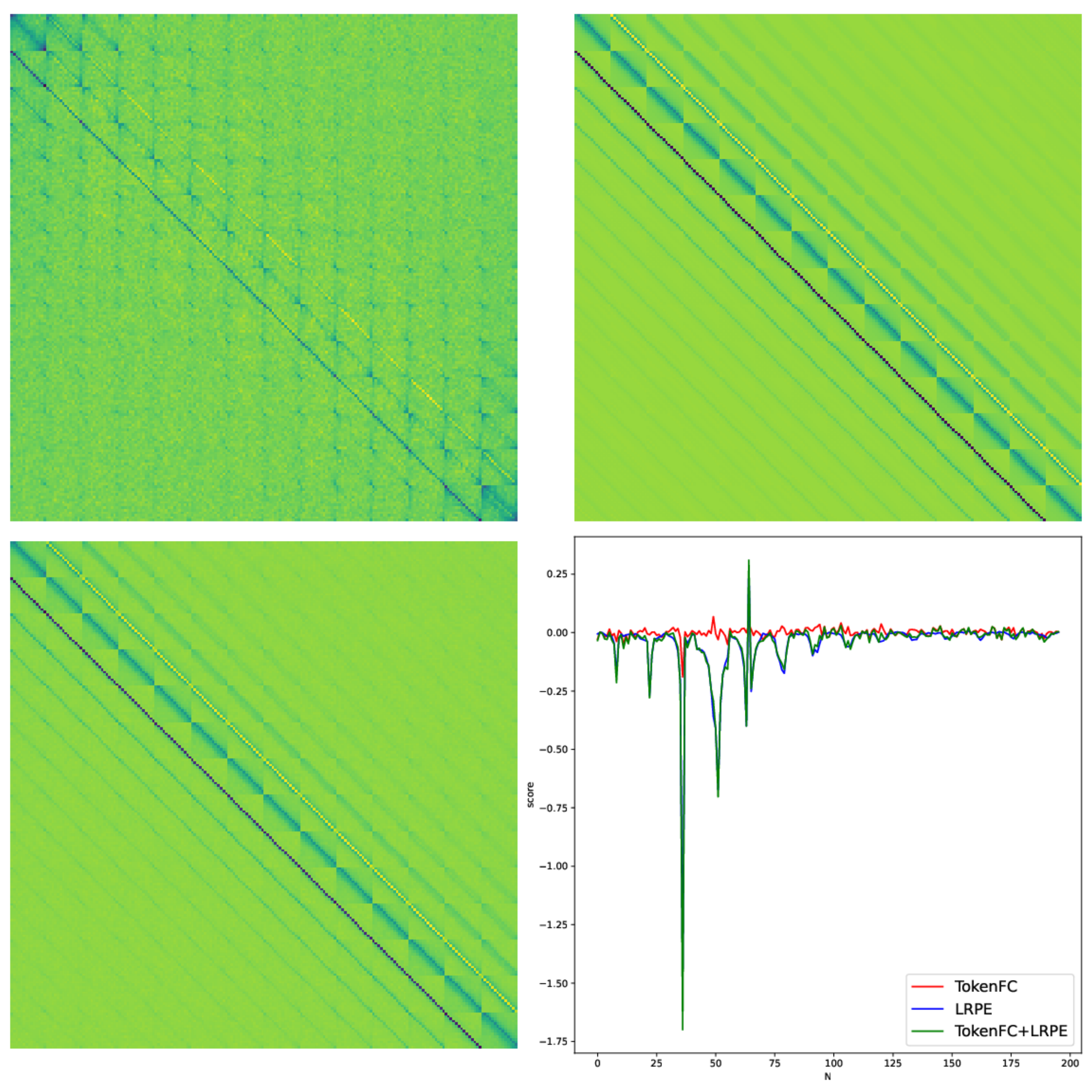}

    \caption{Layer1}
    \label{fig:save_img_1}
  \end{subfigure}
  \hfill
  \begin{subfigure}{0.47\linewidth}
  \includegraphics[width=1\linewidth]{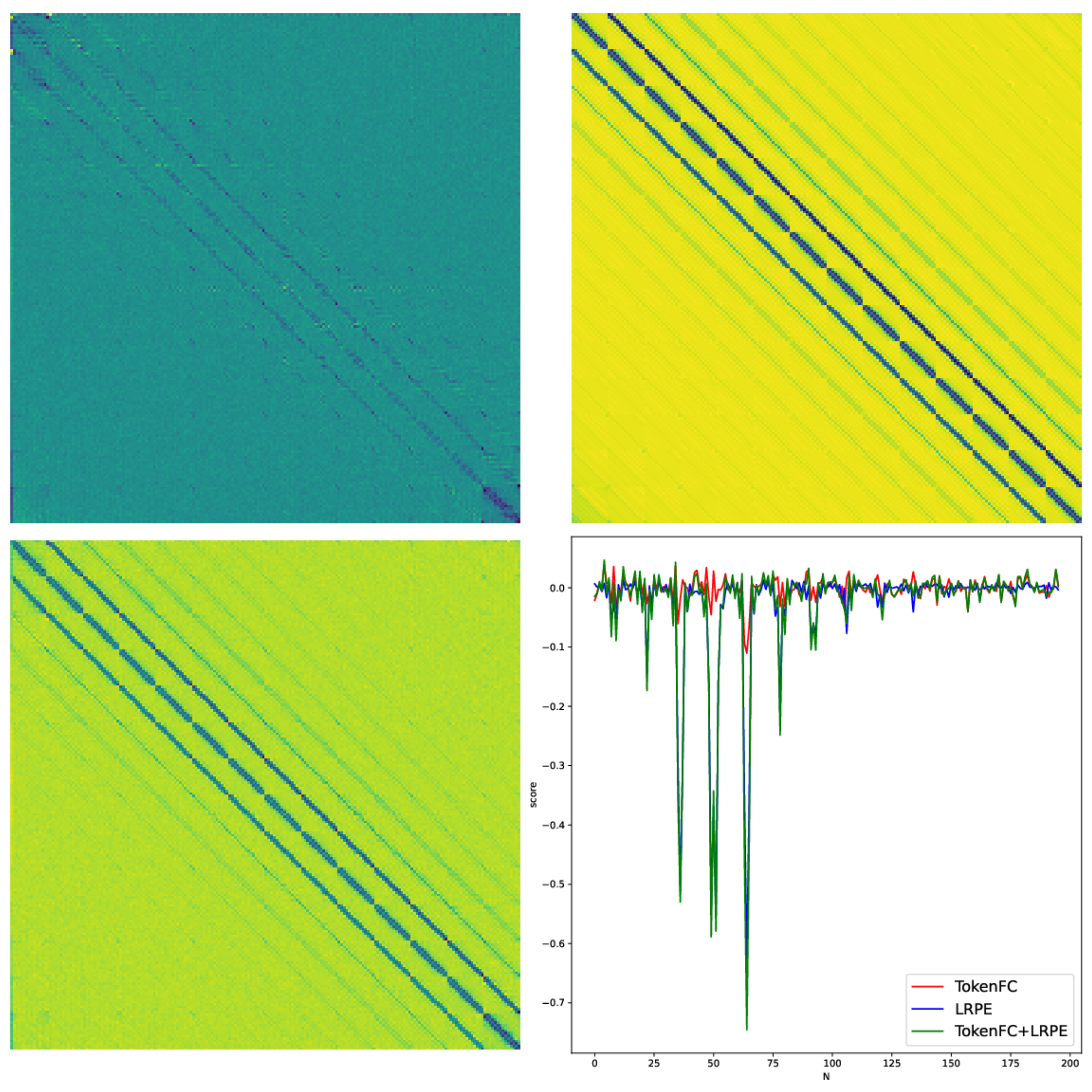}

    \caption{Layer9}
    \label{fig:save_img_7}
  \end{subfigure}
    \hfill
  \begin{subfigure}{0.47\linewidth}  \includegraphics[width=1\linewidth]{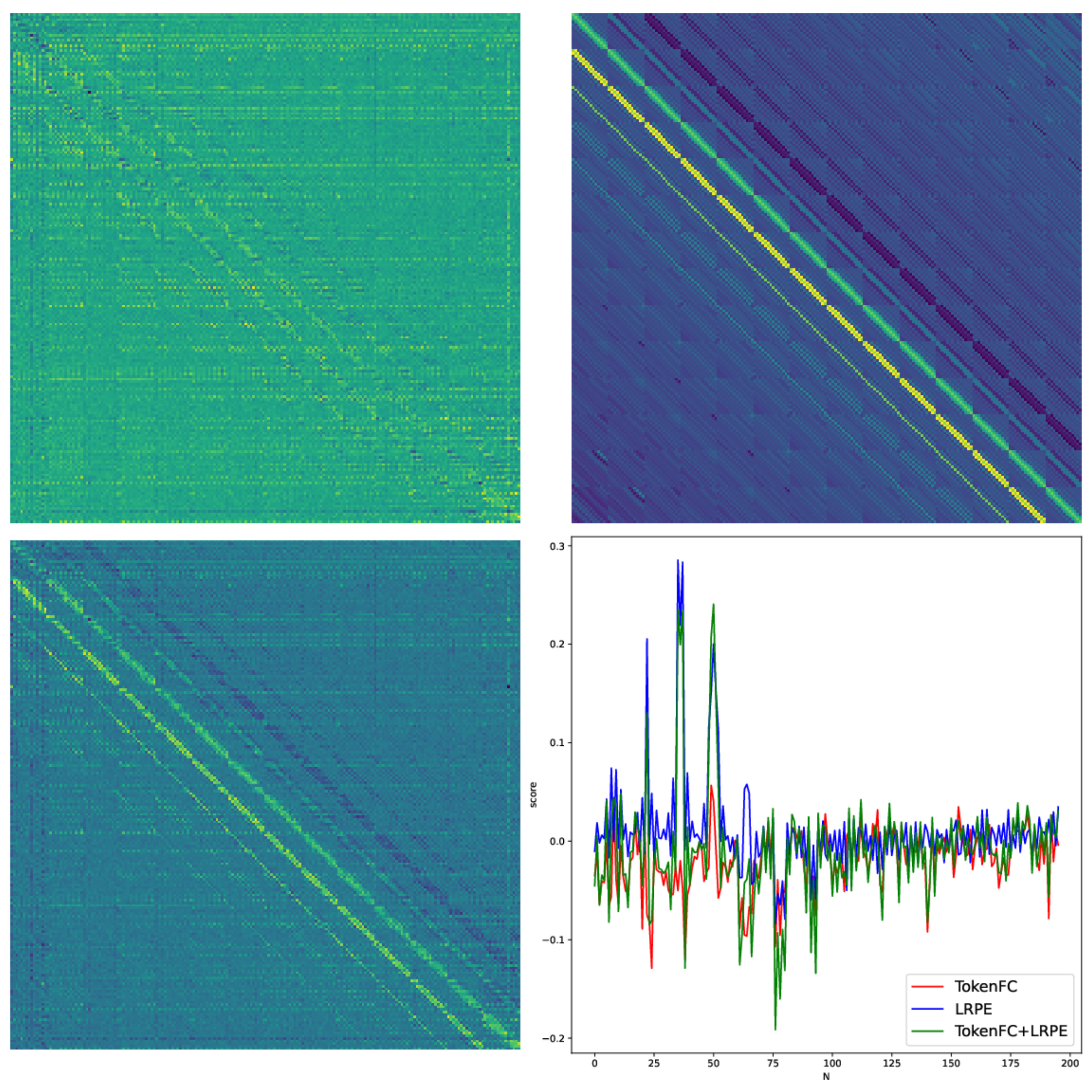}
    \caption{Layer17}
    \label{fig:save_img_18}
  \end{subfigure}
  \hfill
 \begin{subfigure}{0.47\linewidth}
    \includegraphics[width=1\linewidth]{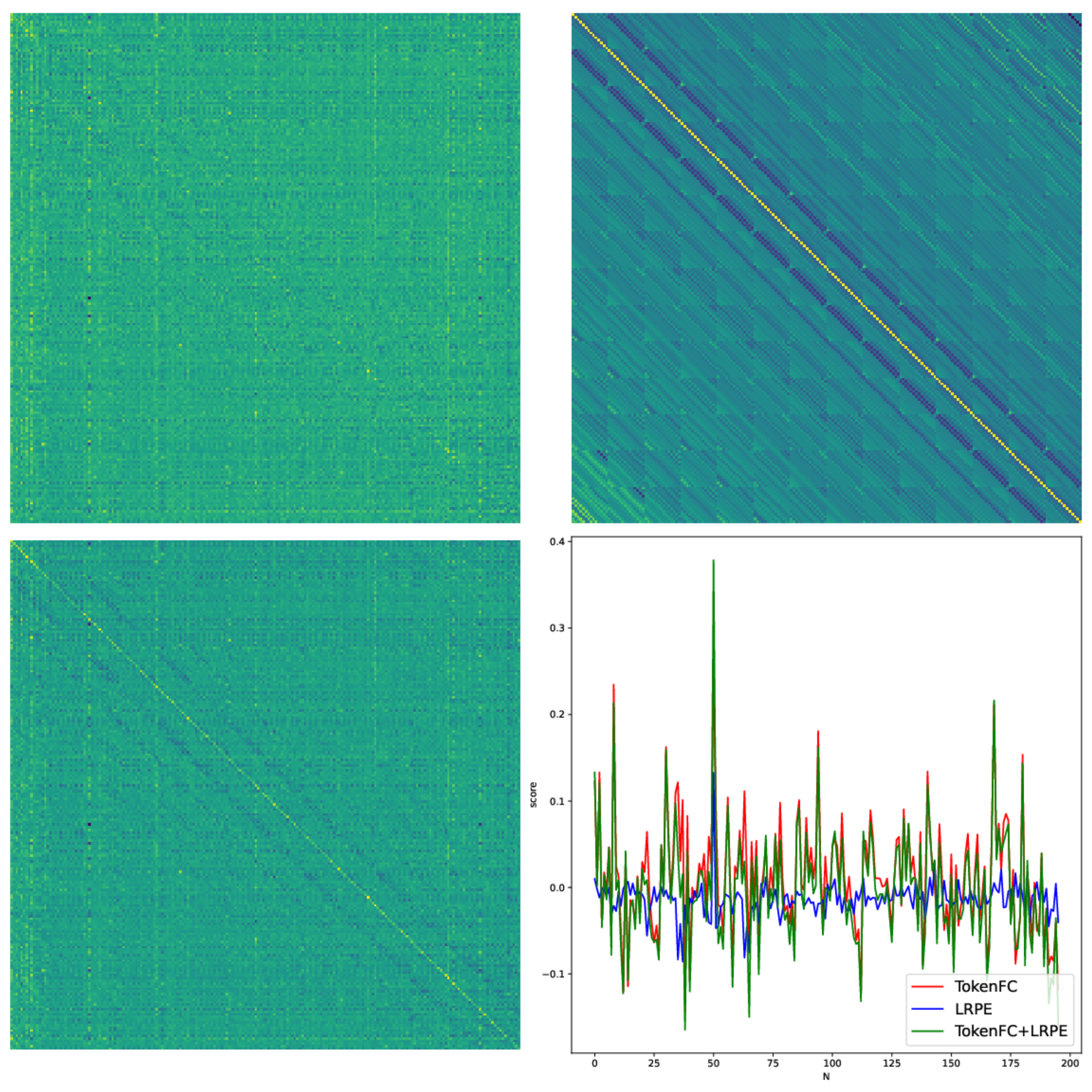}

    \caption{Layer21}
    \label{fig:save_img_23}
  \end{subfigure}
  \hfill
  \caption{\textbf{Visualization of mapping weights after training in Eq. (\ref{eq:9}.)}  Within each subfigure we visualize each component of projection weight matrics of different layers. In each subgraph ,the upper-left is the Token FC's; the upper-right is the LRPE's; The down-left is the TokenFC + LRPE's; The down-right shows the cut line of the row index 40 in each three mapping matrics. The LRPE matrics generally have more regular patterns and keep playing a dominant term in early layers.}
  \label{fig:8}
\end{figure}







\begin{table*}[htb]
  % \small
  \centering
  \renewcommand{\arraystretch}{2}
% 		\resizebox{\linewidth}{24mm}
\setlength{\tabcolsep}{1.4mm}{
  \begin{tabular}{c|c|c|c|c}\hline
     &Ouput size&PosMLP-T& PosMLP-S&PosMLP-B\\ \hline\hline
                  stage 1 & $C\times(56\times56)$ & \makecell[c]{
          $\begin{bmatrix} \text{sz.} 16 \times(14\times14)\\
                             \text{dim} 96, s 8,\gamma 4, \end{bmatrix} \times 2$ }& \makecell[c]{
          $\begin{bmatrix} \text{sz.} 16 \times(14\times14)\\
                             \text{dim} 128, s 8,\gamma 4, \end{bmatrix} \times 2$ }&  \makecell[c]{
          $\begin{bmatrix} \text{sz.} 16 \times(14\times14)\\
                             \text{dim} 192, s 8,\gamma 4, \end{bmatrix} \times 2$ }\\\hline
                  stage 2 & $2C\times(28\times28)$ &\makecell[c]{
          $\begin{bmatrix} \text{sz.} 4 \times(14\times14)\\
                             \text{dim} 192, s 16,\gamma4, \end{bmatrix} \times 2$ }& \makecell[c]{
          $\begin{bmatrix} \text{sz.} 4 \times(14\times14)\\
                             \text{dim} 256, s 16,\gamma 4, \end{bmatrix} \times 2$ }& \makecell[c]{
          $\begin{bmatrix} \text{sz.} 4 \times(14\times14)\\
                             \text{dim} 384, s 16,\gamma4, \end{bmatrix} \times 2$ }\\ \hline        
                  stage 3 & $4C\times(14\times14)$ &\makecell[c]{
          $\begin{bmatrix} \text{sz.} 1 \times(14\times14)\\
                             \text{dim} 384, s 32,\gamma 4, \end{bmatrix} \times 18$ }& \makecell[c]{
          $\begin{bmatrix} \text{sz.} 1 \times(14\times14)\\
                             \text{dim} 512, s 32,\gamma 4, \end{bmatrix} \times 18$ }& \makecell[c]{
          $\begin{bmatrix} \text{sz.} 1 \times(14\times14)\\
                             \text{dim} 768, s 32,\gamma4, \end{bmatrix} \times 18$ }\\ \hline
                   stage 4 & $8C\times(7\times7)$ &\makecell[c]{
          $\begin{bmatrix} \text{sz.} 1 \times(7\times7)\\
                             \text{dim} 768, s 64,\gamma 2, \end{bmatrix} \times 2$ }& \makecell[c]{
          $\begin{bmatrix} \text{sz.} 1 \times(7\times7)\\
                             \text{dim} 1024, s 64,\gamma 2, \end{bmatrix} \times 2$ }& \makecell[c]{
          $\begin{bmatrix} \text{sz.} 1 \times(7\times7)\\
                             \text{dim} 1536, s 64,\gamma 2, \end{bmatrix} \times 2$ }\\\hline
          % Param.& \diagbox[dir=SW]{}{}&18.9M&20.9M&36.8M&82.0M\\\hline
          % FLOPs & \diagbox[dir=SW]{}{}&5.3G&5.2G&8.8G&18.7G\\\hline
  \end{tabular}}

  \caption{Detailed architecture specifications of PosMLP.}
  \vspace{-0.2cm}
  \label{tab:arkitecture}
\end{table*}

\subsection{Non-Localilty}
    
  \begin{figure}[tp]
    \centering
  \begin{subfigure}{0.48\linewidth}
  \centering
  \includegraphics[width=1\linewidth]{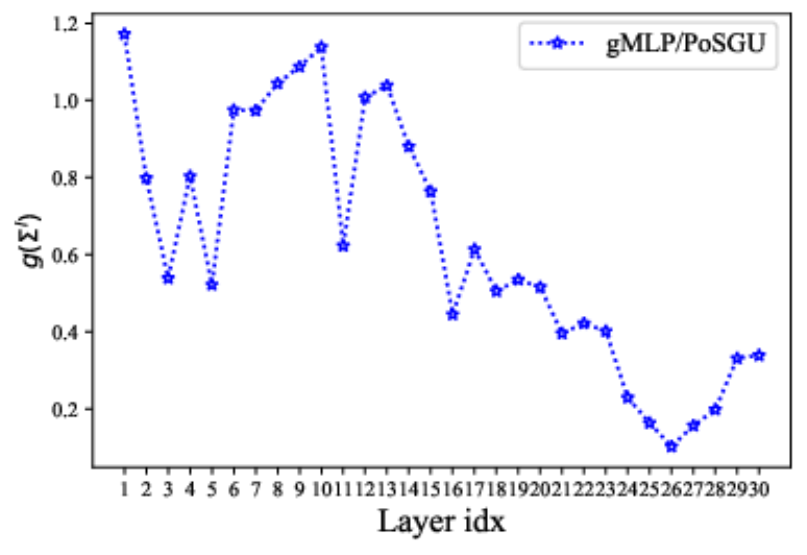}
    \caption{gMLP/PoSGU} 
    \end{subfigure}
  \begin{subfigure}{0.48\linewidth}
  \centering
  \includegraphics[width=1\linewidth]{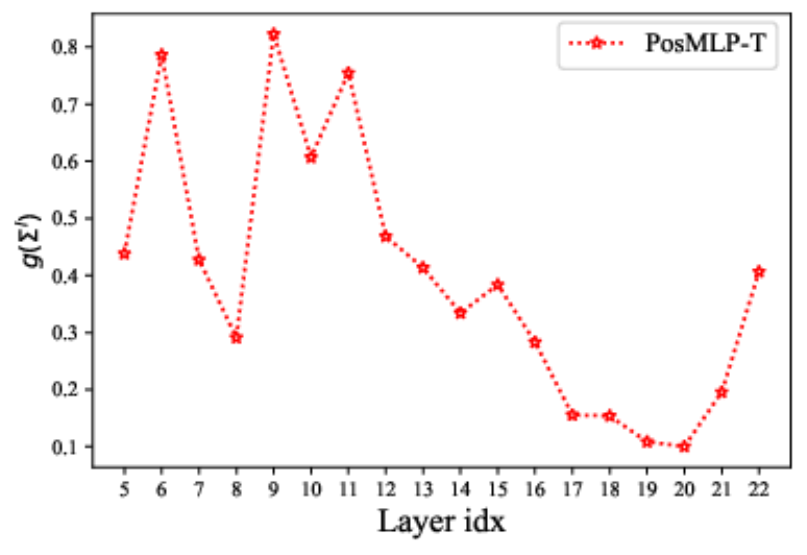}
    \caption{PosMLP/PoSGU} 
    \end{subfigure}
    \vspace{-0.2cm}
    \caption{Degree of non-local property in each PoSGU layer. The non-locality is measured by the designed matrix $g(\tilde{\Sigma}^l)$ where $\mathbf{\Sigma}$ is the covariance matrix. The smaller of $g(\tilde{\Sigma}^l)$ indicates this layer tends to capture more non-local patterns.}
        \vspace{-0.3cm}
    \label{fig:5}
  \end{figure}
% From Zhicai: This part I am not sure the reasonability of designed matrix and statement. Maybe I could put some analysis on Appendix.
Since a $\mathbf{\Sigma}$ determines the logits distribution, it can reveal the non-locality property of the model. Typically the smaller eigenvalues of $\mathbf{\Sigma}$ will help query token attend more to other tokens. Thus we define a matrix to quantitatively visualize the non-localilty as follow:

% As discussed by Cordonnier et al \cite{Onthe}, some $\mathbf{\Sigma}$ tend to be singular or close to $\mathbf{0}$, i.e. some eigenvalue of $\mathbf{\Sigma}$ is extremely small.   
% Here we quantitatively visualize the two terms respectively with naive but demonstrative functions. for a given layer $l$:

\vspace{-0.3cm}
  \begin{align}
    % &h(\bm \Delta^l) = \frac{1}{s}\sum\limits_{i=1}^s||\bm \Delta^l_{i,:}||_2 \\
    &g(\mathbf{\Sigma}^l) = \frac{1}{s}\sum\limits_{i=1}^s\sqrt{\prod\limits_{j=1}^2\mathbf{\lambda}^l_{i,j}}.
  \end{align}
  Where, $\lambda $ is the eigenvalue of $\mathbf{\Sigma}$ and $l$ indicates the layer index. As discussed by Cordonnier \textit{et al} \cite{Onthe}, some $\mathbf{\Sigma}$ tend to be singular or close to $\mathbf{0}$, i.e., some $\lambda $ are extremely small. We do not take those $\mathbf{\Sigma}$ into account and visualize the results of gMLP/PoSGU and PoSMLP ($3th$ stage) in Figure \ref{fig:5}. The deeper layer tends to have smaller $g(\mathbf{\Sigma}^l)$ which indicates the stronger capability of modeling non-locality. Though the deepest layers slightly become more local, it is consistent with the observation in ConViT \cite{Convit} that the localilty is not monotonically decreasing. 
\subsection{Model Complexity Analysis}
\label{Asec:1}
In this part, we give the parameter and computational complexities of the PosMLP block (Single window) and gMLP block. For notation clarity, we denote: input tensor map size as $(H,W,d)$; token number $N = H\times W$; channel expansion ratio in PosMLP block as $\gamma$; group number is denoted as $s$. As such, the parameters number of FC layers in the PosMLP block and gMLP block can be calculated as follows:
\begin{align*}
  P(gMLP)&=\frac{3}{2}\gamma d^2 +(\gamma + 1) d+ N^2+ N , \\
  P(PosMLP/GLRPE)&=\frac{3}{2}\gamma d^2 + (\gamma + 1) d + (4s+1) N - 4s\sqrt{N}+ 4s, \\
  P(PosMLP/GGRPE)&=\frac{3}{2}\gamma d^2 + (\gamma + 1) d + N + 6s.
\end{align*}
The parameters modeling token-mixing significantly shrinks (i.e., $N^2\rightarrow6s$). Their computational complexities are:
\begin{align*}
  \Omega(gMLP)&= \frac{3}{2}\gamma d^2N+\frac{1}{2}\gamma dN^2,\\
  \Omega(PosMLP/GLRPE)&=\frac{3}{2}\gamma d^2N+\frac{1}{2}\gamma dN^2+sN^2,\\
  \Omega(PosMLP/GGRPE)&=\frac{3}{2}\gamma d^2N+\frac{1}{2}\gamma dN^2+ 5sN^2.
    % + \frac{1}{2}h^2w^2c
\end{align*}
Compared with the first two terms, the extra term $5sN^2$ is typically small and bearable. 
\end{document}